\documentclass{article} 
\usepackage{iclr2026_conference,times}

\usepackage{amsmath,amsfonts,bm}

\def\eqref#1{equation~\ref{#1}}

\def\1{\bm{1}}

\DeclareMathAlphabet{\mathsfit}{\encodingdefault}{\sfdefault}{m}{sl}
\SetMathAlphabet{\mathsfit}{bold}{\encodingdefault}{\sfdefault}{bx}{n}

\usepackage{hyperref}
\usepackage{url}
\usepackage{graphicx}
\usepackage{xspace}
\usepackage{multirow} 
\usepackage{booktabs}
\usepackage{amsmath}
\usepackage{enumitem}
\usepackage{wrapfig} 
\usepackage{booktabs} 
\usepackage{pifont}   
\usepackage{array}    
\usepackage{makecell} 

\usepackage{pifont}
\usepackage{xcolor}

\newcommand{\cmark}{\textcolor{green!60!black}{\ding{51}}} 
\newcommand{\xmark}{\textcolor{red}{\ding{55}}}    
\newcommand{\na}{\textcolor{gray}{\textbf{N/A}}}           

\definecolor{bittersweet}{rgb}{1.0, 0.44, 0.37}
\newcommand{\name}{GUI-360$^\circ$\xspace}
\newcommand{\agent}{TrajAgent\xspace}

\title{\name: A Comprehensive Dataset and \\ Benchmark for Computer-Using Agents}

\author{
Jian Mu\textsuperscript{1}, 
Chaoyun Zhang\textsuperscript{2}\thanks{Corresponding author.}, 
Chiming Ni\textsuperscript{3}, 
Lu Wang\textsuperscript{2},  
Bo Qiao\textsuperscript{2},   
Kartik Mathur\textsuperscript{2}, 
\AND
Qianhui Wu\textsuperscript{2}, 
Yuhang Xie\textsuperscript{4}, 
Xiaojun Ma\textsuperscript{2}, 
Mengyu Zhou\textsuperscript{2}, 
Si Qin\textsuperscript{2}, 
Liqun Li\textsuperscript{2}, 
\AND
Yu Kang\textsuperscript{2}, 
Minghua Ma\textsuperscript{2}, 
Qingwei Lin\textsuperscript{2}, 
Saravan Rajmohan\textsuperscript{2}, 
Dongmei Zhang\textsuperscript{2} \\\\
\textsuperscript{1}Nanjing University \quad
\textsuperscript{2}Microsoft \quad
\textsuperscript{3}ZJU-UIUC \quad
\textsuperscript{4}Peking University
}

\iclrfinalcopy 
\begin{document}

\maketitle

\begin{abstract}
We introduce \name, a large-scale, comprehensive dataset and benchmark suite designed to advance \emph{computer-using agents} (CUAs). CUAs present unique challenges and is constrained by three persistent gaps: a scarcity of real-world CUA tasks, the lack of automated collection-and-annotation pipelines for multi-modal trajectories, and the absence of a unified benchmark that jointly evaluates GUI grounding, screen parsing, and action prediction.

\name addresses these gaps with an LLM-augmented, largely automated pipeline for query sourcing, environment-template construction, task instantiation, batched execution, and LLM-driven quality filtering. The released corpus contains over 1.2M executed action steps across thousands of trajectories in popular Windows office applications, and includes full-resolution screenshots, accessibility metadata when available, instantiated goals, intermediate reasoning traces, and both successful and failed action trajectories. The dataset supports three canonical tasks, GUI grounding, screen parsing, and action prediction, and a hybrid GUI+API action space that reflects modern agent designs. Benchmarking state-of-the-art vision--language models on \name reveals substantial out-of-the-box shortcomings in grounding and action prediction; supervised fine-tuning and reinforcement learning yield significant gains but do not close the gap to human-level reliability. We release \name and accompanying code to facilitate reproducible research and accelerate progress on robust desktop CUAs.

The full dataset has been made public on 
\textcolor{bittersweet}{\url{https://huggingface.co/datasets/vyokky/GUI-360}}.
\end{abstract}

\section{Introduction}
Recent advances in vision--language and large language models have sparked rapid progress toward intelligent agents that automate tasks inside digital environments \cite{zhang2024large}. Such agents interpret natural-language requests, perceive screen content via pixels and/or accessibility (a11y) metadata, plan sequences of operations, and then either navigate the GUI or invoke APIs to complete tasks on a user's behalf \cite{zhang2025api}. They can dramatically reduce user effort for routine productivity tasks and enable novel human–computer workflows. However, realizing this potential requires two tightly coupled capabilities: reliable screen understanding (element grounding or screen parsing) \cite{cheng2024seeclick, lu2024omniparser, zheng2025vem} and robust action planning (stepwise action prediction and execution) \cite{zhang2024ufo}. Both capabilities in turn depend critically on large, diverse, and high-quality datasets grounded in realistic execution contexts \cite{wang2024large}.

We focus on a concrete, under-served class of agents  called \emph{computer-using agents} (CUAs) \cite{cua2025}: agents whose primary operating domain is the desktop computer environment. Desktop CUAs differ from web \cite{zheng2024gpt} or mobile \cite{wang2024mobile} agents in several important ways. Desktop applications present very high-resolution mixed-content screens, heterogeneous widgets and document formats, arbitrary window layouts (multi-window and multi-monitor settings), and frequently lack standardized accessibility metadata \cite{zhang2025ufo2}. Tasks on desktop systems are also often longer-horizon and more compositionally structured (e.g., find a table in a document, transform cells in Excel, then copy results into PowerPoint). These characteristics make desktop CUAs substantially more challenging to train and evaluate than their web or mobile counterparts \cite{zhang2024large}.

Despite growing interest, progress on desktop CUAs is hampered by three persistent gaps. First, there is a scarcity of real-world task collections: existing datasets are often handcrafted or LLM-synthesized \cite{sun2024genesis}, which limits their ability to capture the frequency and diversity of authentic user intents. Second, automated pipelines for data collection and annotation are largely missing \cite{nayak2025ui}. Manual execution and labeling of desktop interactions is expensive, error-prone, and difficult to scale, making it impractical to generate multi-modal execution trajectories at scale. Third, no unified, large-scale benchmark exists that supports the breadth of tasks needed for comprehensive evaluation \cite{nayak2025ui, li2025screenspot}. Prior datasets typically focus on a single aspect—such as element detection, a single modality, or a narrow application subset, rather than jointly enabling \emph{GUI grounding}, \emph{screen parsing}, and \emph{action prediction} with execution traces and failure cases.

\begin{figure}[t]
\vspace{-1em}
  \includegraphics[width=\columnwidth]{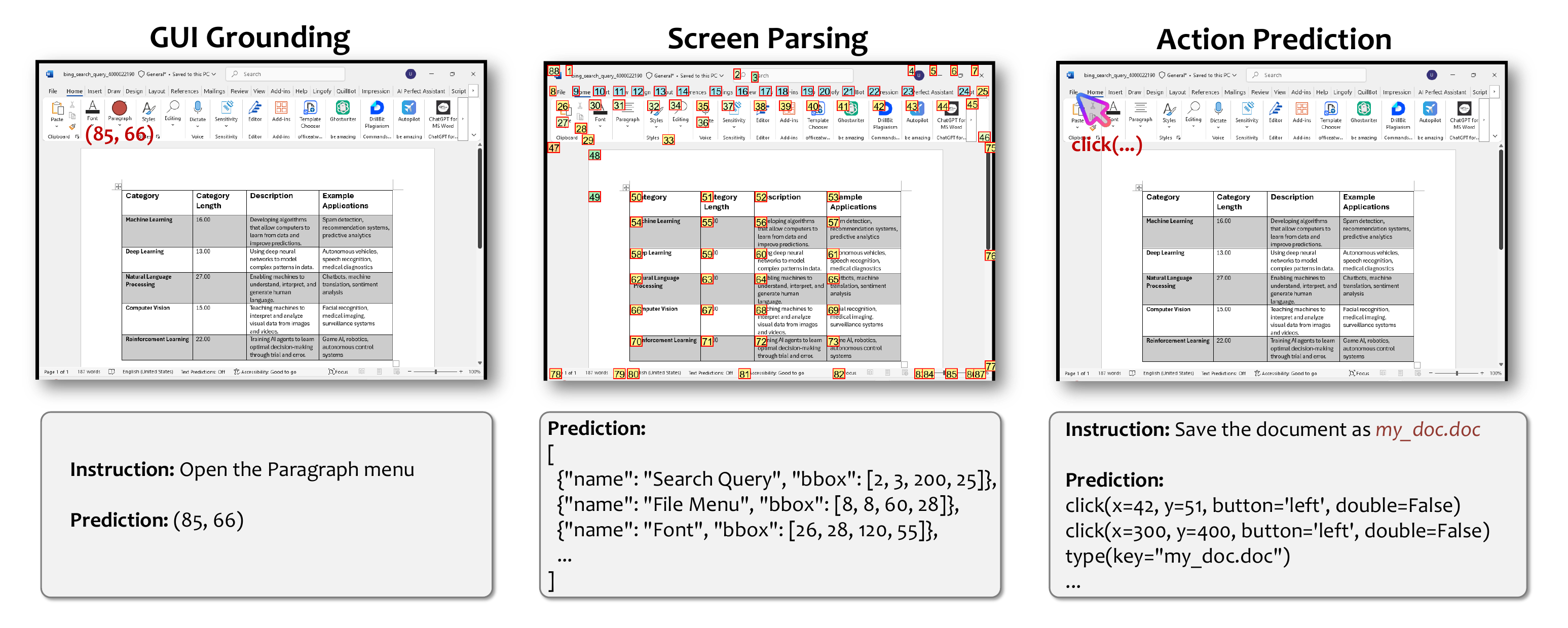}
  \vspace{-2em}
  \caption{An illustration of GUI Grounding, Screen Parsing and Action Prediction tasks included in our \name.}
  \label{fig:tasks}
\end{figure}

To fill these gaps, we introduce \name, a comprehensive dataset and benchmark suite for desktop computer-using agents. \name is built around three design goals: realism, scalability, and task breadth. Concretely, the dataset and benchmark provide the following properties:  
\begin{enumerate}[leftmargin=*]
  \item \textbf{Real-world, high-frequency queries.} Task intents are harvested from authentic sources (search logs, community forums, and in-app help content) to reflect common user needs and high-frequency workflows.  
  \item \textbf{Automated collection and annotation pipeline.} We present an LLM-augmented, largely automated pipeline for template construction, task instantiation, batched execution, and LLM-driven quality filtering, minimizing human intervention while preserving execution realism.  
  \item \textbf{Comprehensive multi-modal annotations.} Each example contains full-resolution screenshots, accessibility metadata, natural-language goals, intermediate agent ``thoughts,'' and stepwise action trajectories (including successful and failed runs). These assets jointly support three canonical tasks: (a) \emph{GUI Grounding}: given a step-level plan, predict the screen coordinates or UI element to interact with; (b) \emph{Screen Parsing}: given a screenshot, enumerate the set of interactable UI elements and their properties; and (c) \emph{Action Prediction}: given the current state and user intent, predict the next action (click/type/select/API call).  We show an illustration of included three tasks in Figure~\ref{fig:tasks}.
  \item \textbf{Large scale and practical coverage.} \name contains over 1.2M executed action steps spanning thousands of trajectories across widely used Windows office applications (Word, Excel, PowerPoint), and is designed so that the template set and pipeline can be extended to other desktop applications.  
  \item \textbf{Hybrid GUI+API action space.} To reflect modern CUA architectures \cite{zhang2025ufo2, zhang2025api}, our action space mixes direct GUI operations with higher-level API calls where available, enabling evaluation of both perception-driven and API-assisted strategies.  
\end{enumerate}
\begin{table*}[t]
\centering
\caption{Comparison of GUI datasets across dimensions. A checkmark (\cmark) indicates support, while a cross (\xmark) indicates not supported.}
\label{tab:gui_datasets}
\renewcommand{\arraystretch}{1.2}
\setlength{\tabcolsep}{4pt}
\resizebox{\columnwidth}{!}{ 
\begin{tabular}{
    l  
    c 
    c c c  
    c 
    c 
    c c  
    c 
    p{1cm}   
    p{1cm} 
}
\toprule
Dataset & \makecell{Query \\ Source} & \multicolumn{3}{c}{Task} & Samples & \makecell{Data \\ Collection} & \multicolumn{2}{c}{Action} & Reasoning & A11y Info. & Fail Case \\
\cmidrule(lr){4-6} \cmidrule(lr){9-10}
 &  &  Grounding & Parsing & Action Pred. &  &  & GUI & API &  &  &  \\
\midrule
ScreenSpot  & \makecell{Human-\\designed} & \cmark & \xmark & \xmark & 1,200+ & Human & \na & \na & \xmark & \xmark & \xmark \\\midrule
ScreenSpot-Pro  & \makecell{Human-\\designed} & \cmark & \xmark & \xmark & 1{,}581 & Human & \na & \na & \na & \na & \na \\\midrule
DeskVision  & Online & \cmark & \xmark & \xmark & 54{,}855 & Auto. & \na & \na & \xmark & \xmark & \xmark \\\midrule
UI-Vision  & \makecell{Human-\\designed } & \cmark & \xmark & \cmark & 8{,}227 & Human & \cmark & \xmark & \xmark & \xmark & \xmark \\\midrule\midrule
\textbf{\name}  & \textbf{\makecell{In-App/Online/\\Search}} & \cmark & \cmark & \cmark & \textbf{1{,}225{,}177} & Auto. & \cmark & \cmark & \cmark & \cmark & \cmark \\
\bottomrule
\end{tabular}
}
\end{table*}
Table~\ref{tab:gui_datasets} compares \name with existing GUI datasets across key dimensions. Unlike prior efforts that focus narrowly on grounding or small-scale scripted tasks, \name provides full coverage of grounding, parsing, and action prediction with a large-scale, automatically collected corpus. Notably, it is the first dataset to include accessibility information, reasoning supervision, and both GUI- and API-level actions, making it a uniquely comprehensive benchmark for CUA research.  

To assess \name’s utility, we benchmark state-of-the-art vision–language models, asking two questions: (i) how well do existing models generalize to realistic desktop CUAs without adaptation, and (ii) how much can fine-tuning on \name bridge the gap? Our results reveal consistent patterns: off-the-shelf models struggle with grounding in heterogeneous layouts and often fail in stepwise action prediction, leading to cascading errors. Training on \name yields significant gains. These findings underscore both the limitations of current models and the value of \name as a scalable, challenging benchmark for driving progress in CUAs.

\section{Related Work}

\paragraph{GUI and Computer-Using Agents}
LLMs have enabled agents \cite{zhang2024large} that automate tasks across web \cite{zheng2024gpt, zheng2025skillweaver}, mobile \cite{wang2024mobile, wang2024mobilev2, wang2025mobile}, and desktop platforms \cite{zhang2024ufo, zhang2025ufo2, qin2025ui}. Desktop environments are particularly challenging for the CUAs due to high-resolution displays and complex layouts. Such agent typically rely on accessibility metadata or visual screen understanding \cite{gur2023real, xie2023openagents}. When accessibility information is missing, they resort to screen parsing  and visual grounding, as in SeeClick \cite{cheng2024seeclick}, OmniParser \cite{lu2024omniparser}, and GUI-Actor \cite{wu2025gui}. UFO \cite{zhang2024ufo, zhang2025ufo2} pioneered hybrid approaches that combine accessibility with screen parsing for more reliable control detection, while recent efforts further integrate APIs for efficiency \cite{zhang2025api}.

Ultimately, progress depends on large-scale data to support robust screen understanding and tool use. \name aims to fill this gap by providing a comprehensive dataset and benchmark for GUI grounding, action prediction, and screen parsing.

\paragraph{Data and Benchmarks for CUA}
A key obstacle in building effective CUAS lies in obtaining high-quality training data and reliable benchmarks. While such resources are increasingly available for web \cite{deng2023mind2web, zhou2023webarena} and mobile domains \cite{rawles2023androidinthewild, rawles2024androidworld}, the desktop setting remains comparatively underexplored despite its greater complexity. Several efforts have emerged in this space. UI-Vision \cite{nayak2025ui} provides a human-annotated desktop benchmark with bounding boxes, UI labels, and action trajectories. DeskVision \cite{xu2025deskvision} introduces a cross-OS dataset for desktop region captioning. OfficeBench \cite{wang2024officebench} offers a live testing environment for office applications through handcrafted cases and oracle evaluation. However, these datasets and benchmarks either require costly human annotation of trajectories and GUI states, or cover only narrow subsets of tasks, limiting their scalability and diversity.

Our work, \name, addresses this gap by introducing an automated data collection pipeline that generates large-scale resources for GUI grounding, screen parsing, and action prediction. This design makes \name the most comprehensive and scalable dataset–benchmark suite to date for training and evaluating GUI agents.

\section{The Collection of \name}
\label{sec:method}

\begin{figure}[t]
\vspace{-1em}
  \includegraphics[width=\columnwidth]{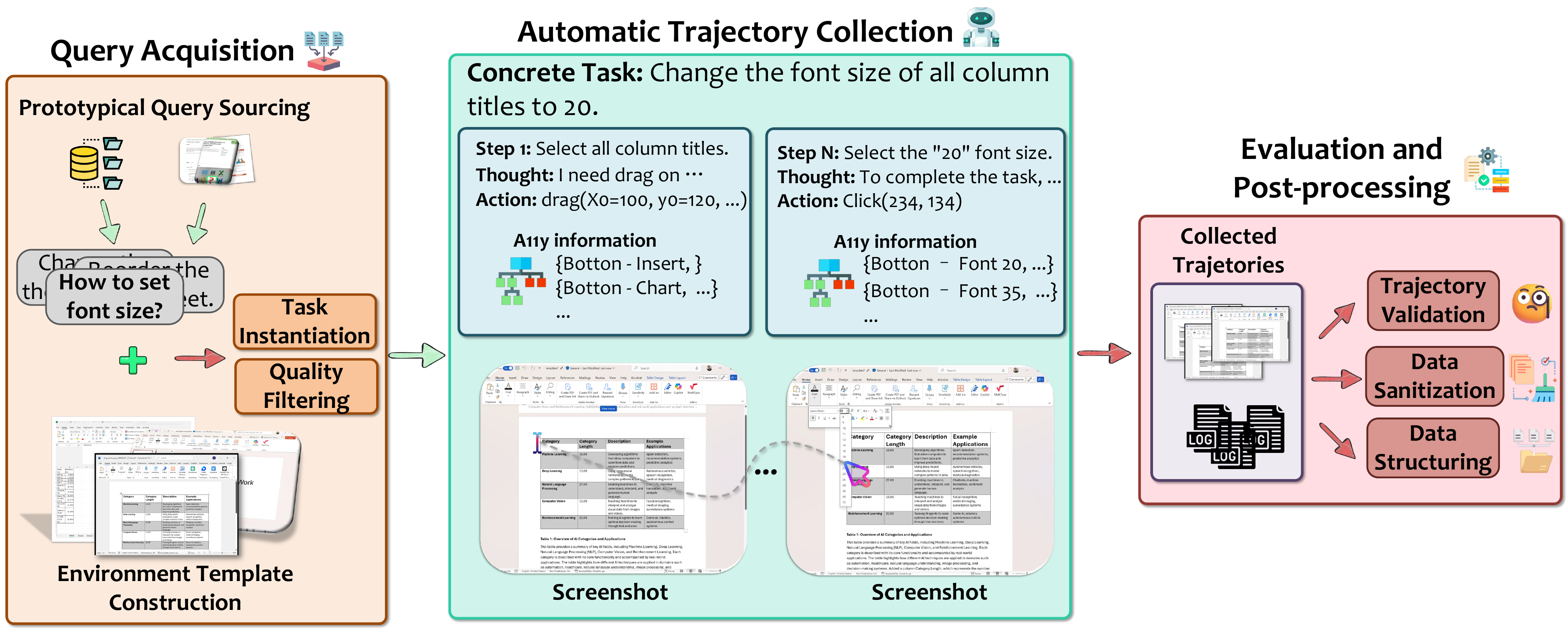}
  \vspace{-2em}
  \caption{The data collection pipeline for \name.}
  \label{fig:pipeline}
\end{figure}

The construction of \name follows a three-stage pipeline designed to maximize scalability while minimizing human effort, as shown in Figure~\ref{fig:pipeline}.
\begin{enumerate}[leftmargin=*]
    \item \textbf{Query Acquisition.} We begin by collecting \emph{real-world} user queries from reliable sources and augmenting them with synthetic variants to ensure coverage of diverse and realistic task intents.
    \item \textbf{Automatic Trajectory Collection.} We design a specialized CUA named \agent to execute tasks in an automatic, consistent and high-quality manner. The \agent generates and collect detailed trajectories that jointly support GUI grounding, screen parsing, and action prediction, enabling multi-task supervision from a single execution.
    \item \textbf{Evaluation and Post-processing.} Finally, we apply automated evaluators and systematic post-processing to verify correctness, filter noise, and enhance overall data quality.
\end{enumerate}
Together, these stages yield a scalable and comprehensive pipeline for constructing \name. By integrating real-world task diversity, automated trajectory collection, and rigorous quality control, our approach provides a unified dataset and benchmark that supports multiple core capabilities of GUI agents.

\subsection{Query Acquisition}
High-quality and actionable user queries are the foundation of a reliable dataset, as they determine both task realism and execution fidelity \cite{xu2025deskvision}. To capture queries that faithfully reflect real-world user needs, we design a dedicated four-stage acquisition pipeline (Figure~\ref{fig:tasks}):
\textit{(i) \textbf{Prototypical Query Sourcing.}} We gather prototypical, high-frequency queries about software usage from diverse and trustworthy real-world sources, ensuring coverage of practical user intents.
\textit{(ii) \textbf{Environment Template Construction.}} Based on the collected queries, we design template software environments that can realistically support their execution. These templates encode the minimal yet sufficient context needed for task instantiation.
\textit{(iii) \textbf{Task Instantiation.}} Each query is grounded into a suitable environment by matching it to the best-fitting template. We then rephrase the query into a concrete, executable form tailored to the selected environment.
\textit{(iv) \textbf{Quality Filtering.}} Finally, we apply post-processing to discard low-quality or ambiguous queries, retaining only those that are actionable and fully grounded in executable environments.
This four-stage pipeline ensures that the resulting queries are both realistic and executable, providing a strong basis for collecting high-quality trajectories in later stages of the \name pipeline.

\paragraph{Prototypical Query Sourcing.}

\begin{table}[t]
\centering
\caption{Raw query statistics across applications and sources.}
\label{tab:raw_query_stats}
\begin{tabular}{l c c c}
\toprule
\textbf{Source} & \textbf{Word} & \textbf{Excel} & \textbf{PowerPoint} \\
\midrule
In-App  & 274   & 159   & 316   \\
Online  & 1,914 & 3,393 & 1,701 \\
Search  & 25,715 & 25,000 & 19,681 \\
\midrule
Total   & 27,903 & 28,552 & 21,698 \\
\bottomrule
\end{tabular}
\end{table}

Most existing CUA datasets rely heavily on human-crafted instructions or LLM-generated queries, which often fail to reflect real-world usage patterns or capture high-frequency tasks. To better ground our dataset in authentic user behavior, we source prototypical task descriptions from three complementary channels:
\begin{enumerate}[leftmargin=*]       
    \item \textbf{In-App Help Content (In-APP)}: We mine built-in help documentation and tutorials of software applications, which provide standardized, well-structured task descriptions covering core functionalities.
    \item \textbf{Online Websites (Online)}: We crawl forums, Q\&A platforms, and community-driven websites to obtain diverse and authentic task descriptions contributed by real users.
    \item \textbf{Search Queries (Search)}:  We extract queries from search engines that mention or relate to the target software. These queries capture pressing, high-frequency user needs and common practical challenges. 
\end{enumerate}
We show the queries distribution from different source in Table~\ref{tab:raw_query_stats}. By combining these sources, we ensure that the collected queries balance authenticity (from search and community data) with completeness (from in-app documentation), resulting in a broad and realistic coverage of user intents. 


\paragraph{Environment Template Construction.}
Each user query must be grounded in a suitable environment that provides the necessary context for agent execution. For example, a task such as ``make the first line of text bold'' in Word requires a document containing editable text; without such a setup, the query is not actionable. Naively creating a bespoke environment for every query would be prohibitively expensive and redundant, since many queries share common contextual requirements.

To address this, we introduce an environment template construction process that systematically amortizes environment setup across queries. Specifically, we leverage GPT to analyze each query and extract its underlying requirements (e.g., the presence of text, a table, or an image). Queries with similar requirements are clustered and abstracted into environment template descriptions, which specify the minimal context needed for execution. We then manually instantiate a curated set of high-frequency templates from these descriptions. 

In practice, we design 30 templates for Word, 30 for Excel, and 6 for PowerPoint. Despite this relatively \textbf{small set of 66} templates, the coverage is substantial: a single template can accommodate diverse content and scenarios, enabling the collection to support approximately \textbf{95\%} of prototypical queries. This template-driven strategy dramatically reduces human effort by avoiding per-query environment creation, and ensure consistent and reproducible environments, enabling robust execution and trajectory collection at scale.

\paragraph{Task Instantiation.} 
Prototypical queries collected from real-world sources are often vague and underspecified (e.g., \textit{``How to make text bold?''}). To be executable by an agent, each query must be grounded in a specific environment and reformulated into an actionable instruction (e.g., \textit{``Make the phrase `Hello World' bold in the Word document''}). We refer to this process as \emph{task instantiation}. To systematically achieve this, we design an automated two-stage pipeline:  
\begin{enumerate}[leftmargin=*]
    \item \textbf{Template Matching.} For each query, we retrieve candidate environment templates. Each template is defined by a set of contextual constraints (e.g., a Word document containing text, or an Excel sheet with numerical entries). Using an LLM, we provide the query together with textual descriptions and screenshots of all available templates. The LLM identifies the most suitable template by reasoning over query requirements and environment affordances. Queries that fail to match any template are discarded.  
    \item \textbf{Query Concretization.} Once a template is selected, the LLM rephrases the vague query into a fully instantiated one, grounded in the chosen environment. For example, the generic request \textit{``make text bold''} is concretized as \textit{``make the text `Hello World' bold in the current document''}. This ensures the query is unambiguous, executable, and directly tied to a well-defined environment.  
\end{enumerate}  
This pipeline is fully automated and enforces that all instantiated tasks are (i) \emph{actionable}, by guaranteeing an associated environment exists, (ii) \emph{concrete}, by eliminating ambiguity in the query, and (iii) \emph{scalable}, as the process can be applied to thousands of queries with minimal human intervention. In practice, this approach yields a large collection of executable tasks while maintaining high fidelity to real-world user intent.  

\paragraph{Quality Filtering.} 
Although task instantiation produces a large set of grounded queries, not all of them are suitable for reliable agent training. Many instantiated tasks may suffer from contextual mismatches, external dependencies, or inherent ambiguities. To ensure robustness, we design a \emph{task filtering pipeline} that employs an LLM as an automatic quality gate. Concretely, the LLM-based judge evaluates each candidate task against a set of well-defined constraints. Tasks are discarded if they:  
\begin{enumerate}[leftmargin=*]
    \item \textbf{Non-Executable (\texttt{NONEXEC})}:  
    Inputs that do not describe a concrete action are removed. This includes subjective statements, vague preferences, or general inquiries that cannot be directly executed. For instance, instructions containing words such as ``custom,'' ``you want,'' or undefined operations like ``edit text'' without specifying the target element fall under this category. 

    \item \textbf{Cross-Application Dependency (\texttt{CROSSAPP})}:  
    Tasks that require interaction with applications beyond \{app\} are excluded. This includes operations that involve opening or manipulating content in external software (e.g., Excel, Edge, File Explorer, or system settings). Representative examples include merging files across applications, printing documents (which requires printer integration), or exporting data to a third-party tool. 

    \item \textbf{Version Management (\texttt{VERCTRL})}:  
    Tasks that involve checking, updating, downgrading, or modifying the version of \{app\} are discarded. Since version management depends heavily on system environment and external factors, these tasks are not considered executable within the scope of our benchmark. 

    \item \textbf{Template Dependency (\texttt{TPLMISS})}:  
    Tasks that rely on specific document or workspace templates that are absent from the provided context are excluded. For example, instructions that assume the existence of a predefined table, chart, or object not available in the given file are categorized as \texttt{TPLMISS}. Note that application-wide settings (e.g., enabling dark mode) do not fall under this restriction, as they do not depend on document templates. 

    \item \textbf{Irrelevant or Invalid (\texttt{INVALID})}:  
    Remaining cases that do not fit into the above categories, or are otherwise infeasible due to irrelevance, ambiguity, or context mismatch, are marked as invalid. For example, a task unrelated to \{app\} or lacking sufficient contextual information for execution would be discarded under this category. 
\end{enumerate}

\begin{table}[t]
\centering
\caption{Distribution of task categories after filtering, shown as percentages of total candidate tasks for each application.}
\label{tab:task_filtering_stats}
\begin{tabular}{lccc}
\toprule
\textbf{Category} & \textbf{Word} & \textbf{Excel} & \textbf{PowerPoint} \\
\midrule
\texttt{CROSSAPP} & 15.61\% & 13.27\% & 12.13\% \\
\texttt{TPLMISS}  &  2.93\% &  4.18\% &  7.51\% \\
\texttt{NONEXEC}  &  1.56\% &  1.68\% &  3.65\% \\
\texttt{VERCTRL}  &  0.17\% &  0.13\% &  0.27\% \\
\texttt{INVALID}  &  2.70\% &  6.98\% &  1.46\% \\\hline\hline
\texttt{NORMAL}   & 77.02\% & 73.76\% & 74.97\% \\ 
\bottomrule
\end{tabular}
\end{table}

Table~\ref{tab:task_filtering_stats} summarizes the outcome of our LLM-based task filtering across Word, Excel, and PowerPoint. Overall, we retain 59,553 out of 79,075 candidate tasks (75.3\%) as \texttt{NORMAL} (self-contained, concretely specified, and template-feasible); the remaining 24.7\% are filtered for various reasons.


\noindent\textbf{Summary.}  
The complete query acquisition pipeline transforms raw, vague user queries into a curated collection of high-quality, executable tasks. Starting from diverse sources of authentic user requests, we progressively ground them into concrete environments through template construction, instantiate them into actionable forms, and finally enforce strict quality filtering. This systematic process yields a dataset that is both scalable and faithful to real-world user intent, while maintaining the rigor necessary for robust agent training.

\subsection{Automatic Trajectory Collection}
Once queries have been instantiated and grounded, the next step is to generate corresponding action plans, execution trajectories, and outcomes. Traditionally, this stage relies on human operators to perform tasks and log interactions, which is costly and difficult to scale.  To overcome this limitation, we develop a specialized execution framework, \agent, that automatically completes queries in batch (\ref{fig:pipeline}) and records detailed execution data trajectories. The workflow consists of three main phases:
\begin{enumerate}[leftmargin=*]
    \item \textbf{Environment Preparation.} For each task, \agent initializes the corresponding environment template and ensures all preconditions for execution are satisfied.  
    \item \textbf{Task Execution and Data Logging.} The agent performs the query step-by-step, generating a complete action trajectory while recording the full GUI state at each step. Additional relevant data, such as intermediate screenshots, control properties, and execution metadata, are also captured.  
    \item \textbf{Environment Reset.} After execution, \agent closes or resets the environment to a clean state, preparing it for the next task in the batch.  
\end{enumerate}
This fully automated process ensures that every task is executed reliably, trajectories are captured consistently, and human intervention is entirely eliminated. By integrating environment preparation, execution, and logging into a single pipeline, \agent enables large-scale, high-fidelity trajectory collection suitable for training and benchmarking CUAs.

\subsubsection{\agent Design}
\begin{figure}[t]
\vspace{-1em}
  \includegraphics[width=\columnwidth]{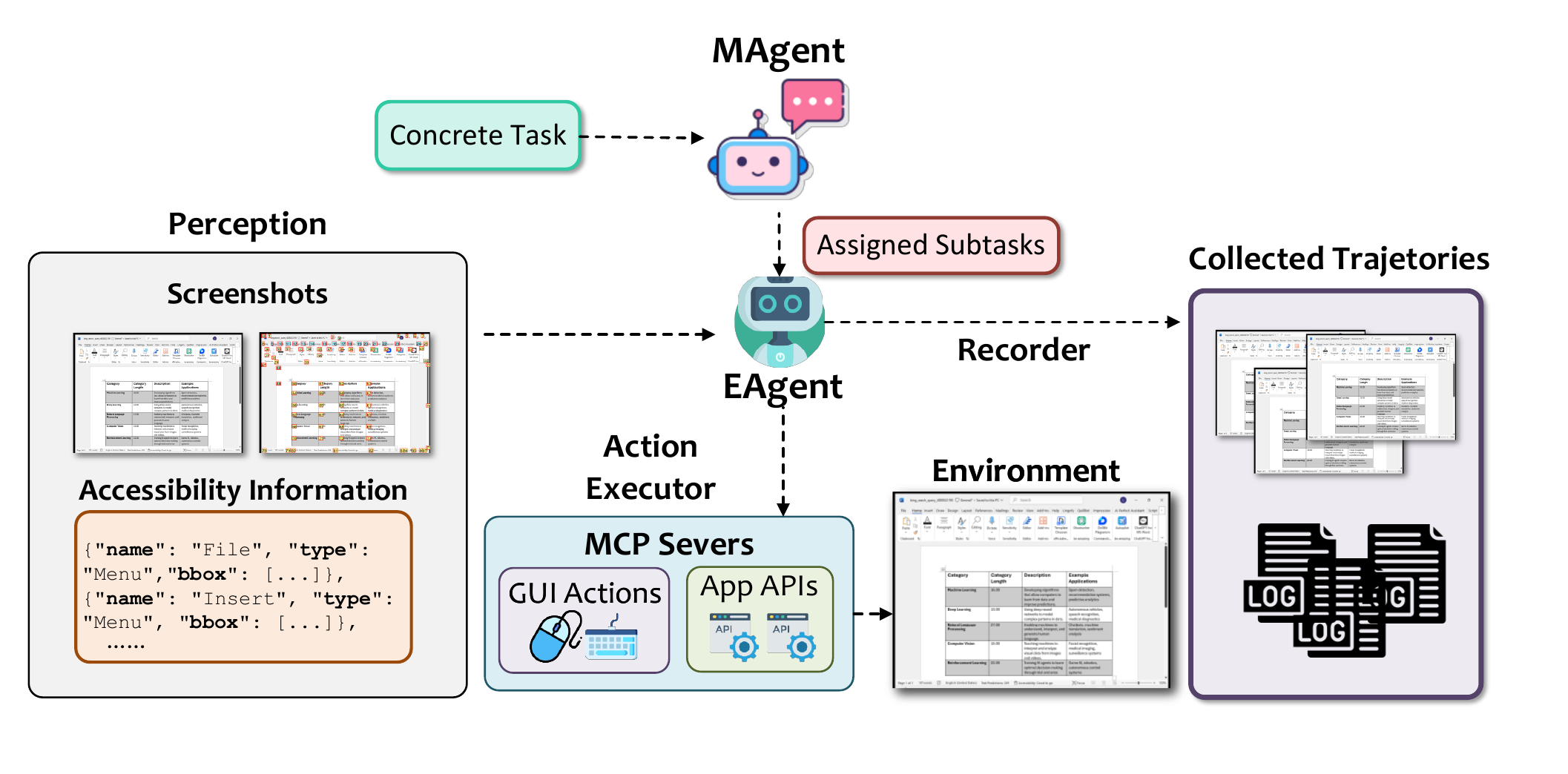}
  \vspace{-4em}
  \caption{The overall architecture of the \agent.}
  \label{fig:trajagent}
\end{figure}
The core requirement for large-scale trajectory collection is twofold: the executor must \textit{(i)} complete instantiated queries with a high success rate to maximize data efficiency, and \textit{(ii)} record every element required for dataset construction (screen snapshots, accessibility metadata, action logs, pre/post states, etc) with strict fidelity. To meet these requirements we design \agent, an orchestrated, multi-agent execution framework that reliably completes concrete tasks and produces high-fidelity execution traces suitable for downstream training and evaluation.

\paragraph{Architecture overview.} \agent follows an orchestration pattern \cite{zhang2024ufo} composed of a \texttt{MasterAgent} (MAgent), a pool of \texttt{ExecutionAgents} (EAgent), with a set of auxiliary services (Perception, Action Executor, and Recorder), as shown in Figure~\ref{fig:trajagent}. The MAgent receives an concrete task and decomposes it into a sequence of manageable subtasks (planning). Each subtask is dispatched to an available EAgent for execution. EAgents operate as lightweight workers that (a) perceive the current GUI state, (b) select or synthesize the next low-level action (click/type/select/API call), (c) execute the action via UI automation or API invocation, and (d) return observations and status back to the MAgent. The Recorder persistently logs the full GUI state and metadata before and after each action, ensuring a temporally-aligned dataset.

\paragraph{Perception.}
Within each EAgent, the Perception service captures a full-resolution screenshot at every decision step and queries the Windows accessibility API (UI Automation) \cite{haverty2005new} to extract a list of actionable controls (name, type and exact bounding box). The accessibility-derived control metadata is rendered on the screenshot as a Set-of-Mark (SoM) \cite{yang2023set}. The EAgent uses both the raw screenshot and the SoM/controls for decision-making: screenshots provide visual context while accessibility metadata supplies precise semantic and locational information, reducing visual reasoning overhead.

\paragraph{Action Executor.}
The Action Executor utilizes app-specific MCP servers \cite{hou2025model} to provide an extensible set of tools for the EAgent. In addition to conventional GUI actions (e.g., mouse clicks, keyboard input), our design incorporates app-level API actions, reflecting modern CUA practices. These API actions improve task efficiency and serve as reliable fallbacks when GUI interactions are prone to failure. At each step, the EAgent selects the most appropriate action based on its current observation, internal reasoning, and task plan, following the classical ReAct paradigm \cite{yao2023react}. This iterative perception–reasoning–action loop continues until the task is fully completed, ensuring both robustness and fidelity in trajectory collection.

\paragraph{Recorder.}
The Recorder is responsible for collecting all multi-modal information at each execution step to construct the dataset. This includes screenshots, accessibility metadata, and agent outputs. Importantly, a single execution of the agent produces data for all downstream tasks, significantly improving data efficiency. Table~\ref{tab:task_io} summarizes the input and output for each task type. Accessibility-derived data ensures precise element locations and properties, while screenshots provide full visual context.  

\paragraph{Two-Stage Execution.}
\begin{table}[t]
\centering
\caption{Success rate of the two-stage execution strategy across applications.}
\label{tab:two_stage_execution}
\begin{tabular}{lcccc}
\toprule
 & Word & Excel & PowerPoint & Total \\
\midrule
Round 1 (GPT-4o) & 16.65\% & 9.27\% & 8.00\% & 11.63\% \\
Round 2 (GPT-4.1) & 16.03\% & 18.20\% & 14.91\% & 16.38\% \\
\midrule
Overall & 30.50\% & 25.78\% & 21.71\% & 26.09\% \\
\bottomrule
\end{tabular}
\end{table}
Leveraging the components described above, queries are executed in batches within a Windows Sandbox, with the EAgent automatically collecting all required data. To reduce dependency on the capabilities of a single model and improve coverage, we adopt a two-stage execution strategy. In the first stage, GPT-4o serves as the base model to complete the queries. Any queries that \textbf{fail in this stage} are then re-executed using a stronger model GPT-4.1. As shown in Table~\ref{tab:two_stage_execution}, the two-stage execution strategy substantially improves success rates compared to relying on a single model. In the first round, GPT-4o captures a non-trivial portion of tasks, particularly in Word. In the second round, GPT-4.1 recovers many of the failures from GPT-4o, with notable gains in Excel and PowerPoint. Together, this staged approach boosts overall completion to 26.09\%, showing that cascading models of complementary strengths increases both success rate and dataset versatility, while avoiding over-reliance on a single model.

\begin{table}[t]
\centering
\caption{Task input-output specifications for dataset collection.}
\label{tab:task_io}
\begin{tabular}{l|p{4cm}|p{6cm}}
\toprule
\textbf{Task} & \textbf{Input} & \textbf{Output} \\
\midrule
GUI Grounding & Application screenshot, Agent's thought at the current step & Operation coordinates of the target element, obtained via accessibility APIs \\
\midrule
Screen Parsing & Application screenshot & List of all actionable controls on screen with name and bounding box, e.g., \texttt{\{"name": "Open Menu", "bbox": [12,34,56,78]\}} \\
\midrule
Action Prediction & User query, Application screenshot, Accessibility information (optional) & Action call, with optional metadata such as agent's thought and plan \\
\bottomrule
\end{tabular}
\end{table}

\subsection{Evaluation and Post-processing}
\label{sec:post-process}
To ensure the high quality and usability of the automatically collected trajectories, we perform a three-stage post-processing procedure:  
\textit{(i)} \textbf{Trajectory Validation.} Each trajectory is automatically evaluated to retain only successful and executable task executions, ensuring that downstream training and evaluation are based on realistic completions.  
\textit{(ii)} \textbf{Data Sanitization.} Low-quality steps, incomplete records, or any data that fail to meet predefined quality criteria are removed. This step eliminates noise and increases the overall reliability of the dataset.  
\textit{(iii)} \textbf{Data Structuring.} The cleaned trajectories are reformatted and normalized into the required structure for the 3 downstream tasks. This includes standardizing screenshot metadata, accessibility information, action calls, and annotations to create a consistent, machine-readable dataset.  

Together, these stages guarantee that the final dataset is both high-quality and fully compatible with diverse CUA training and benchmarking pipelines.

\paragraph{Trajectory Validation.}
To ensure the reliability of collected data, each trajectory undergoes automatic validation. 
We design an evaluation agent, \textit{EvaAgent}, which leverages GPT-4.1 in an LLM-as-a-judge paradigm \cite{gu2024survey}.  EvaAgent inspects the trajectory step by step, including the screenshot, accessibility information, executed actions, intermediate thoughts, and the final application state. Following prior work~\cite{wang2024large}, it employs a chain-of-thought style reasoning process to decompose the query into several fine-grained evaluation criteria. A trajectory is marked as successful only if all criteria are satisfied, thereby enforcing a stricter notion of task completion. 

To assess its reliability, we conducted a small-scale study on 100 randomly sampled trajectories. EvaAgent's judgments achieved 86\% agreement with human annotators, demonstrating that it provides sufficiently accurate and scalable validation. Compared to hard-coded scripts or brittle oracle rules, this approach offers greater flexibility across diverse applications and enables rapid filtering of high-quality, successful trajectories at scale.

\paragraph{Data Sanitization.}
After validation, we perform a final cleaning step to ensure completeness and consistency of the collected trajectories. 
This involves removing any step that lacks an executed action, a screenshot, or essential metadata required for downstream tasks.  Such sanitization further improves the overall data quality and ensures that only fully executable, well-documented steps are retained for model training and evaluation.

\paragraph{Data Structuring.}
Finally, we transform the sanitized data into a standardized JSON format tailored for model consumption, following the input–output specifications summarized in Table~\ref{tab:task_io}. 
For the \emph{Action Prediction} task, we provide two input modalities: \textit{visual-only} and \textit{visual+a11y}:
\begin{itemize}[leftmargin=*]
    \item \textbf{Visual-only:} The model receives raw screenshots as input. Interaction-related arguments (e.g., click positions) are represented as the absolute coordinates of the center of the corresponding bounding box. 
    \item \textbf{Visual+a11y:} The model additionally receives the list of actionable elements from the accessibility API, which are also annotated on the screenshot using the Set-of-Mark (SoM) representation. Interaction arguments are expressed as element ID and name, chosen from the provided element list. This reduces the need for explicit coordinate prediction and lowers the visual grounding overhead.
\end{itemize}
For evaluation, we partition the trajectories into training (80\%) and benchmark (\textsc{GUI-360}-Bench, 20\%) splits. All three tasks—GUI Grounding, Screen Parsing, and Action Prediction—share the same data partition to maintain consistency across evaluations.

\subsection{\name Statistics}
Following the pipeline described above, we construct a comprehensive dataset, \name, for training across three core GUI tasks, and a companion benchmark, \name-Bench, for systematic evaluation.

\begin{table}[t]
\centering
\caption{Training and test dataset statistics across domains (Word, Excel, PowerPoint).}
\label{tab:train_test_stats}
\begin{tabular}{lrrrr}
\toprule
\multicolumn{5}{c}{\textbf{\name-Train}} \\
\midrule
 & Word & Excel & PowerPoint & Total \\
\midrule
Total Trajectories & 5,633 & 4,348 & 3,769 & 13,750 \\
Total Steps & 41,742 & 29,363 & 34,263 & 105,368 \\
Average Steps per Trajectory & 7.41 & 6.75 & 9.09 & 7.66 \\
Steps for Grounding Tasks & 30,695 & 22,319 & 26,473 & 79,487 \\
Steps for Screen Parsing & 41,742 & 29,363 & 34,263 & 105,368 \\
Steps for Action Prediction & 41,742 & 29,363 & 34,263 & 105,368 \\
Total Elements & 3,270,104 & 12,211,852 & 2,186,738 & 17,668,694 \\
Total Images & 83,484 & 58,726 & 68,526 & 210,736 \\
Average Elements per Image & 78.34 & 415.89 & 63.82 & 167.69 \\
GUI Action Rate (\%) & 76.1 & 80.7 & 87.4 & 81.0 \\
API Action Rate (\%) & 23.9 & 19.3 & 12.6 & 19.0 \\
\midrule
\multicolumn{5}{c}{\textbf{\name-Bench}} \\
\midrule
 & Word & Excel & PowerPoint & Total \\
\midrule
Total Trajectories & 1,409 & 1,087 & 943 & 3,439 \\
Total Steps & 10,597 & 7,175 & 8,512 & 26,284 \\
Average Steps per Trajectory & 7.52 & 6.60 & 9.03 & 7.64 \\
Steps for Grounding Tasks & 7,784 & 5,444 & 6,552 & 19,780 \\
Steps for Screen Parsing & 10,597 & 7,175 & 8,512 & 26,284 \\
Steps for Action Prediction & 10,597 & 7,175 & 8,512 & 26,284 \\
Total Elements & 839,273 & 2,940,016 & 545,328 & 4,324,617 \\
Total Images & 21,194 & 14,350 & 17,024 & 52,568 \\
Average Elements per Image & 79.20 & 409.76 & 64.07 & 164.53 \\
GUI Action Rate (\%) & 76.2 & 80.2 & 87.5 & 81.0 \\
API Action Rate (\%) & 23.8 & 19.8 & 12.5 & 19.0 \\
\bottomrule
\end{tabular}
\end{table}

\paragraph{Scale.}
Table~\ref{tab:train_test_stats} summarizes the statistics of \name-Train (80\%) and \name-Bench (20\%). 
In total, \name contains \textbf{13,750 trajectories} with over \textbf{105k steps}, averaging \textbf{7.66 steps per trajectory}. 
The dataset also provides \textbf{210k screenshots} paired with \textbf{17.7M annotated UI elements}, yielding rich multimodal supervision at both the visual and accessibility levels. For evaluation, \name-Bench adds a further \textbf{3,439 trajectories} and \textbf{26k steps}, maintaining a similar distribution of average step length and GUI/API action rates. In addition, we include 62{,}170 trajectories comprising 1{,}093{,}525 steps for failure cases, which can serve as valuable signals for reinforcement learning, similar to the approach in \cite{wang2024large}. These failure cases capture challenging or error-prone situations that models often struggle with, providing rich supervision for improving robustness and reliability.

This unprecedented scale, both in interaction traces and annotated elements, makes \name one of the largest and most comprehensive resources for GUI learning—sufficiently large to train high-capacity models and to rigorously benchmark their generalization in realistic desktop environments.

\paragraph{Diversity.}
\begin{wrapfigure}{r}{0.5\columnwidth}
  \centering
  \includegraphics[width=0.48\columnwidth]{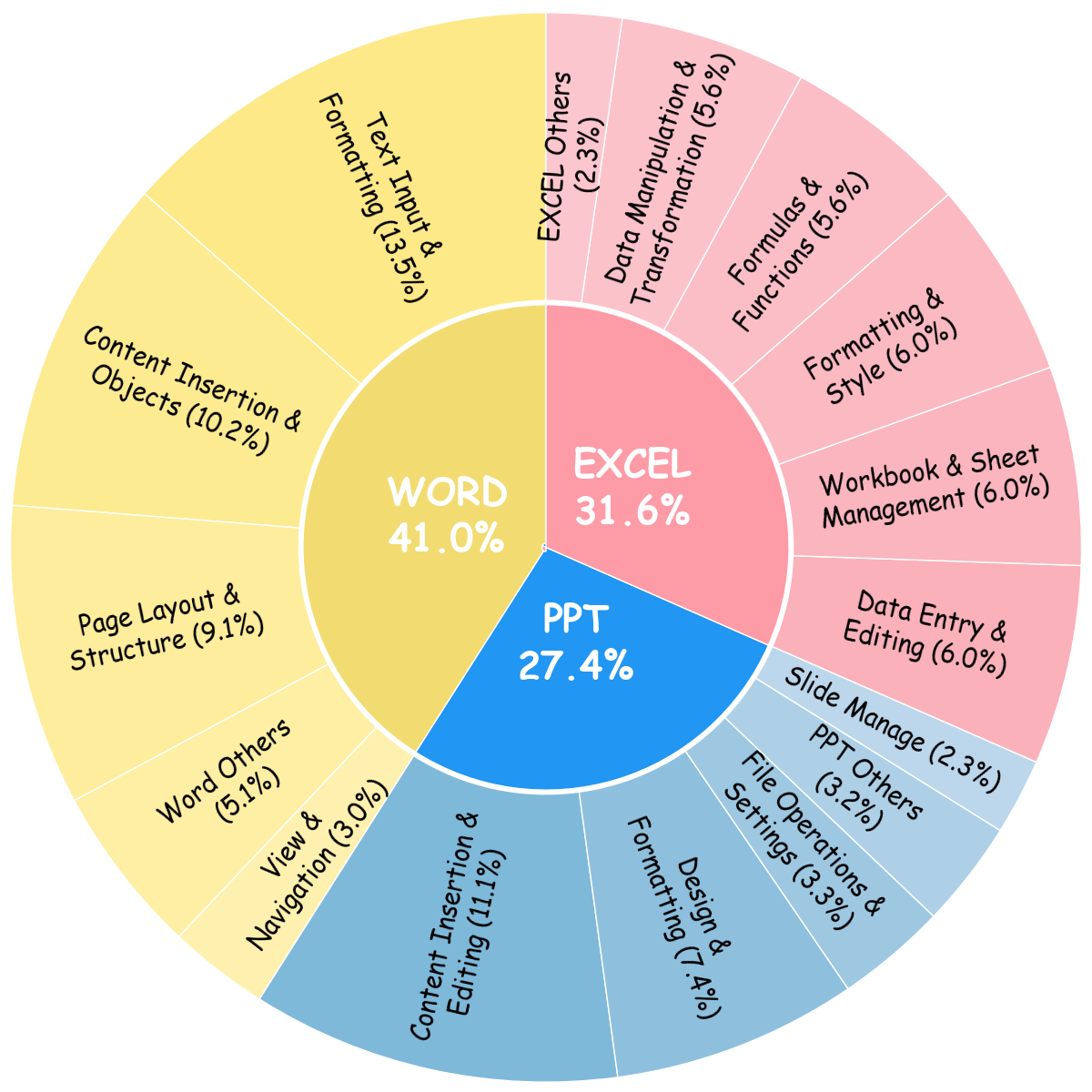}
  \caption{\textbf{Dataset Composition}. For each app, tasks are divided into six categories according to their core operational intent.}
  \label{fig:dataset_distribution}
\end{wrapfigure}
Beyond scale, \name emphasizes breadth and functional diversity. Using GPT-4o, we classified user queries into fine-grained categories, as shown in Figure~\ref{fig:dataset_distribution}. The corpus spans Word (41.0\%), Excel (31.6\%), and PowerPoint (27.4\%), each with rich internal coverage. Word tasks range from text formatting to layout and review, Excel covers data entry, formulas, and visualization, while PowerPoint emphasizes content editing, design, and transitions. This balance ensures exposure to both frequent operations and rarer, long-tail behaviors. As a result, models trained on \name are encouraged to generalize across routine workflows while remaining robust to less common but practically important tasks, making it a comprehensive and challenging benchmark for desktop CUA research.


\section{Experiment}

Our experimental evaluation proceeds in two stages. First, we perform an out-of-the-box evaluation of several state-of-the-art vision--language and agent models on \name-Bench to measure their zero-/few-shot capabilities and to diagnose common failure modes. Second, we investigate how targeted training on \name (supervised fine-tuning and policy optimization) improves model performance and robustness. All experiments use the same data partitioning: the training split described in Section~\ref{sec:post-process} (80\%) and the held-out benchmark \name-Bench (20\%).

\subsection{GUI Grounding}
We begin with the GUI grounding task: given a natural-language task description and the current GUI state, the model must predict the screen location for the next interaction (represented as a 2D coordinate). Predicted coordinates are evaluated against the accessibility-derived bounding box of the target element.

\paragraph{Baselines.}
We evaluate a mix of general-purpose and domain-specialized models. The general-purpose VLM/LLM baselines are GPT-4o \cite{hurst2024gpt}, GPT-4.1 \cite{openai2025gpt4_1}, o3 \cite{openai2025o3_o4mini}, and GPT-5 \cite{openai2025gpt5}. We also include several open-source and grounding-focused models: Qwen-VL-2.5 (7B) \cite{bai2025qwen2}, UGround-7B \cite{gou2024navigating}, Aguvis-7B \cite{xu2024aguvis}, UI-TARS-1.5 (7B) \cite{qin2025ui}, and GUI-Actor (7B) \cite{wu2025gui}. Finally, we report results for supervised fine-tuned variants (Qwen-2.5 7B-SFT, UI-TARS-1.5 7B-SFT) that are trained on the \name training split.

\paragraph{Performance Metrics.}  
The primary evaluation metric for GUI grounding is \textbf{accuracy}, defined as the proportion of predictions where the predicted coordinate \(\hat{c}_i\) lies within the bounding box of the corresponding ground-truth target element \(b_i\). Formally,  
\[
\text{Acc} = \frac{1}{N}\sum_{i=1}^{N} \mathbb{1}\{\hat{c}_i \in b_i\},
\]
where \(N\) is the total number of test cases and \(\mathbb{1}\{\cdot\}\) is the indicator function.

We report accuracy separately for each application domain (\emph{Word}, \emph{Excel}, and \emph{PowerPoint}) as well as an overall aggregated score across all benchmark examples. To highlight the effect of task-specific adaptation, we evaluate two settings:  
(i) the \emph{zero-shot} performance of each baseline model directly on \name-Bench, and  
(ii) the performance after \emph{supervised fine-tuning (SFT)} on the \name{} training set.


\begin{table}[t]
\centering
\caption{Performance of different models on the GUI grounding task across applications.}
\label{tab:gui_grounding}
\begin{tabular}{lcccc}
\toprule
Model & Word & Excel & PowerPoint & Overall \\
\midrule
GPT-4o & 15.22\% & 5.08\% & 5.41\% & 9.38\% \\
GPT-4.1 & 17.30\% & 7.48\% & 7.01\% & 11.41\% \\
o3 & 36.62\% & 19.44\% & 31.06\% & 29.96\% \\
GPT-5 & 34.52\% & 20.31\% & 17.36\% & 25.34\% \\
Qwen-2.5-VL-7B & 38.09\% & 26.76\% & 41.55\% & 35.78\% \\
UGround-7B & 57.44\% & 43.09\% & 59.53\% & 53.85\% \\
Aguvis-7B & 53.14\% & 37.69\% & 59.57\% & 50.50\% \\
UI-TARS-1.5 7B & 63.50\% & 58.61\% & 64.21\% & 62.27\% \\
GUI-Actor 7B & 54.84\% & 45.84\% & 62.68\% & 54.50\% \\\hline
Qwen-2.5-VL-7B-SFT & 84.11\% & 79.20\% & 82.84\% & 82.30\% \\
UI-TARS-1.5 7B-SFT & 84.73\% & 79.84\% & 81.98\% & 82.49\% \\
\bottomrule
\end{tabular}
\end{table}

\paragraph{Performance Comparison.}
Table~\ref{tab:gui_grounding} reports the evaluation results of different models on \name-bench for the GUI grounding task. 
We observe that general-purpose GPT models (e.g., GPT-4o and GPT-4.1) achieve only modest performance, with overall accuracy below 12\%. 
More advanced general models such as GPT-o3 and GPT-5 show improvements (20--30\%), yet still struggle with precise GUI grounding. 
Domain-specific pretraining brings substantial gains: models like UGround-7B and GUI-Actor 7B surpass 50\%, demonstrating the effectiveness of grounding-oriented pretraining. 
Finally, supervised fine-tuning (SFT) on \name yields the largest performance leap, with Qwen-2.5 7B-SFT and UI-TARS-1.5 7B-SFT achieving over 82\% accuracy across applications. 
This progression clearly highlights the value of \name for both training and benchmarking: it not only reveals the limitations of general-purpose models on GUI tasks but also provides high-quality training data that enables fine-tuned models to achieve state-of-the-art performance.

\subsection{Screen Parsing}

The screen parsing task requires a model to take a clean screenshot as input and output the complete set of \emph{interactable} UI elements on the screen. Each predicted element consists of a semantic name (e.g., ``Start Menu'') and a bounding box. This task is challenging due to heterogeneous widgets, dense layouts, occlusions, and mixed content (text, icons, images).

\paragraph{Baselines.}  
We evaluate both general-purpose VLMs and specialized screen-parsing models. General models include GPT-4o~\cite{hurst2024gpt}, GPT-4.1~\cite{openai2025gpt4_1}, GPT-o3~\cite{openai2025o3_o4mini}, GPT-5~\cite{openai2025gpt5}, and Qwen-VL-2.5 (7B)~\cite{bai2025qwen2}. Specialized baselines include OmniParser and OmniParser-v2~\cite{lu2024omniparser}.

\paragraph{Evaluation metrics.}  
We measure parsing quality along three complementary axes: (i) element detection accuracy (precision / recall / F1), (ii) localization quality (mean IoU on matched pairs), and (iii) semantic name accuracy (average text embedding similarity on matched pairs). All metrics are computed per image and then averaged across the benchmark (macro-average).

Let \(G\) be the ground-truth set of elements for an image and \(P\) the predicted set. We obtain a one-to-one matched set \(M \subseteq P \times G\) by performing greedy bipartite matching sorted by descending IoU, and keeping only pairs with \(\mathrm{IoU}>0.5\). For a predicted box \(p\) and ground-truth box \(g\) we use the standard intersection-over-union:
\[
\mathrm{IoU}(p,g) \;=\; \frac{\mathrm{area}(p \cap g)}{\mathrm{area}(p \cup g)}.
\]

For each image \(i\) define:
\[
\text{Precision}_i \;=\; \frac{|M_i|}{|P_i|}, \qquad
\text{Recall}_i \;=\; \frac{|M_i|}{|G_i|}, \qquad
\text{F1}_i \;=\; \frac{2\cdot\text{Precision}_i\cdot\text{Recall}_i}{\text{Precision}_i+\text{Recall}_i},
\]
where \(|\cdot|\) denotes set cardinality. The reported precision, recall, and F1 are the macro-averages across images:
\[
\text{Precision} \;=\; \frac{1}{N}\sum_{i=1}^{N}\text{Precision}_i,\quad
\text{Recall} \;=\; \frac{1}{N}\sum_{i=1}^{N}\text{Recall}_i,\quad
\text{F1} \;=\; \frac{1}{N}\sum_{i=1}^{N}\text{F1}_i.
\]

To quantify localization quality, we compute the mean IoU for each image \(i\) over the matched pairs \(M_i\). If an image has no matched pairs (\(|M_i| = 0\)), we define its IoU as 0. The overall mean IoU is then the macro-average across all images:
\[
\overline{\mathrm{IoU}} \;=\; \frac{1}{N}\sum_{i=1}^{N} 
\begin{cases} 
\frac{1}{|M_i|}\sum_{(p,g)\in M_i} \mathrm{IoU}(p,g), & |M_i|>0 \\[2mm]
0, & |M_i|=0
\end{cases}.
\]

Similarly, for semantic-name accuracy, we embed predicted and ground-truth names using a sentence encoder \(\phi(\cdot)\) (e.g., a sentence-transformer) and compute the cosine similarity for each matched pair. If an image has no matches, we assign a similarity of 0 for that image. The macro-average over all images is:
\[
\overline{\mathrm{Sim}} \;=\; \frac{1}{N}\sum_{i=1}^{N} 
\begin{cases} 
\frac{1}{|M_i|}\sum_{(p,g)\in M_i} 
\frac{\langle \phi(\text{name}_p), \phi(\text{name}_g)\rangle}
{\|\phi(\text{name}_p)\| \, \|\phi(\text{name}_g)\| }, & |M_i|>0 \\[1mm]
0, & |M_i|=0
\end{cases}.
\]

Together, these metrics separate \emph{whether} elements are detected (precision/recall/F1), \emph{how precisely} they are localized (mean IoU), and \emph{how well} their semantic roles are recovered (mean embedding similarity).

\paragraph{Performance Comparison.}
Table~\ref{tab:screen_parsing} presents a detailed comparison of general-purpose VLMs and specialized screen parsing models across Word, Excel, and PowerPoint. Overall, general-purpose models such as GPT-4o, GPT-4.1, GPT-o3, and GPT-5 struggle with both element detection and localization, achieving low F1 scores (0.019–0.128) and moderate mean IoU values (0.229–0.578). Notably, GPT-o3 exhibits the highest overall F1 among the general models (0.128) and maintains relatively strong localization (IoU 0.578), but its recall remains limited, indicating many missed elements. GPT-4.1 and GPT-5 show uneven performance across applications: GPT-5 performs best on Excel (F1 0.126) but poorly on PowerPoint (F1 0.059), suggesting sensitivity to layout complexity and domain-specific content.

In contrast, specialized parsers significantly outperform general-purpose models in all metrics. OmniParser and OmniParser-v2 achieve overall F1 scores above 0.40 and mean IoU above 0.73, with strong text similarity (0.565–0.568), demonstrating robust detection, accurate localization, and reliable semantic recovery. The incremental improvement from OmniParser to OmniParser-v2 is modest but consistent, reflecting refinement in handling dense layouts and occlusions. Application-wise, Word and PowerPoint benefit most from these specialized models due to dense interactive regions, while Excel remains challenging because of its compact grid structure, though performance still exceeds general-purpose VLMs.

These results reveal two key insights: (i) general-purpose VLMs are limited in screen parsing due to the need for fine-grained spatial reasoning and UI semantics, and (ii) task-specific training, as in OmniParser, provides substantial gains in both element coverage and semantic correctness, highlighting the necessity of specialized architectures for accurate and reliable screen understanding.

\begin{table}[t]
\centering
\caption{Comparison of different models across domains (Precision, Recall, F1, Text Similarity, Avg IOU Accuracy).}
\label{tab:screen_parsing}
\begin{tabular}{l l r r r r r}
\toprule
Model & Domain & Precision & Recall & F1 & Text Sim. & Avg IOU \\
\midrule
\multirow{4}{*}{GPT-4o} 
 & Word & 0.040 & 0.017 & 0.024 & 0.170 & 0.252 \\
 & Excel & 0.020 & 0.002 & 0.004 & 0.085 & 0.133 \\
 & PowerPoint & 0.037 & 0.021 & 0.026 & 0.171 & 0.282 \\
 & \textbf{Overall} & 0.034 & 0.014 & 0.019 & 0.147 & 0.229 \\
\midrule
\multirow{4}{*}{GPT-4.1} 
 & Word & 0.101 & 0.065 & 0.077 & 0.307 & 0.518 \\
 & Excel & 0.102 & 0.026 & 0.039 & 0.278 & 0.514 \\
 & PowerPoint & 0.091 & 0.073 & 0.080 & 0.330 & 0.480 \\
 & \textbf{Overall} & 0.098 & 0.057 & 0.067 & 0.306 & 0.505 \\
\midrule
\multirow{4}{*}{o3} 
 & Word & 0.173 & 0.128 & 0.144 & 0.481 & 0.631 \\
 & Excel & 0.178 & 0.099 & 0.118 & 0.335 & 0.523 \\
 & PowerPoint & 0.129 & 0.109 & 0.115 & 0.526 & 0.559 \\
 & \textbf{Overall} & 0.160 & 0.114 & 0.128 & 0.456 & 0.578 \\
\midrule
\multirow{4}{*}{GPT-5} 
 & Word & 0.106 & 0.079 & 0.088 & 0.315 & 0.615 \\
 & Excel & 0.172 & 0.109 & 0.126 & 0.274 & 0.574 \\
 & PowerPoint & 0.065 & 0.056 & 0.059 & 0.314 & 0.508 \\
 & \textbf{Overall} & 0.111 & 0.080 & 0.089 & 0.304 & 0.569 \\
 \midrule
 \multirow{4}{*}{Qwen2.5-VL-7B} 
 & Word & 0.384 & 0.014 & 0.023 & 0.137 & 0.358 \\
 & Excel & 0.041 & 0.002 & 0.003 & 0.082 & 0.094 \\
 & PowerPoint & 0.047 & 0.011 & 0.014 & 0.111 & 0.128 \\
 & \textbf{Overall} & 0.181 & 0.010 & 0.015 & 0.113 & 0.211 \\
\midrule
\multirow{4}{*}{OmniParser} 
 & Word & 0.392 & 0.520 & 0.440 & 0.619 & 0.730 \\
 & Excel & 0.431 & 0.217 & 0.270 & 0.450 & 0.748 \\
 & PowerPoint & 0.417 & 0.588 & 0.479 & 0.594 & 0.718 \\
 & \textbf{Overall} & 0.411 & 0.459 & 0.406 & 0.565 & 0.731 \\
\midrule
\multirow{4}{*}{OmniParser v2} 
 & Word & 0.396 & 0.525 & 0.444 & 0.625 & 0.738 \\
 & Excel & 0.431 & 0.217 & 0.270 & 0.450 & 0.748 \\
 & PowerPoint & 0.418 & 0.590 & 0.481 & 0.596 & 0.721 \\
 & \textbf{Overall} & 0.413 & 0.462 & 0.408 & 0.568 & 0.735 \\
\bottomrule
\end{tabular}
\end{table}

\subsection{Action Prediction}
The action prediction task bridges the gap between a user's natural language command and the executable action calls required by the agent. This represents the ultimate goal of contextual user automation (CUA): translating abstract intent into precise, structured interactions with the GUI. As discussed in Section~\ref{sec:post-process}, we consider two evaluation settings: \emph{visual-only} (where the agent has access solely to screenshots) and \emph{visual+a11y} (where accessibility metadata is also provided). The latter setting is designed to test how much accessibility information improves grounding and execution.

\paragraph{Baselines.}  
We evaluate a set of general-purpose VLMs, including GPT-4o~\cite{hurst2024gpt}, GPT-4.1~\cite{openai2025gpt4_1}, GPT-o3~\cite{openai2025o3_o4mini}, GPT-5~\cite{openai2025gpt5}, and Qwen-VL-2.5 (7B)~\cite{bai2025qwen2}. In addition, we fine-tune Qwen-VL-2.5 (7B) via supervised fine-tuning (SFT) and reinforcement learning (RL) on \name{} to examine post-training improvements.

\paragraph{Performance Metrics.}  
The evaluation of action prediction is more nuanced than grounding, since each action step is composed of a \emph{function}, a set of \emph{arguments}, and a \emph{status} flag (continue or finish). We therefore report three component accuracies and one aggregated metric:  

\begin{itemize}[leftmargin=*]
    \item \textbf{Function accuracy} (\(\text{Acc}_{\text{func}}\)): the proportion of predictions where the predicted function \(\hat{f}_i\) exactly matches the ground-truth function \(f_i\).  
    \[
    \text{Acc}_{\text{func}} = \frac{1}{N}\sum_{i=1}^{N}\mathbb{1}\{\hat{f}_i = f_i\}.
    \]

    \item \textbf{Argument accuracy} (\(\text{Acc}_{\text{args}}\)): evaluated conditionally on the predicted function.  
    If the function is a spatial action such as \texttt{click}, correctness requires that the predicted coordinate \((\hat{x}_i,\hat{y}_i)\) falls inside the ground-truth bounding box \(b_i\).  
    For symbolic arguments (e.g., menu item name, keystroke, value,), correctness requires exact match between predicted and ground-truth arguments.  
    Formally,  
    \[
    \text{Acc}_{\text{args}} = \frac{1}{N}\sum_{i=1}^{N}\mathbb{1}\{\hat{a}_i \equiv a_i \mid f_i\},
    \]
    where the equivalence relation \(\equiv\) depends on the function type \(f_i\).  

    \item \textbf{Status accuracy} (\(\text{Acc}_{\text{status}}\)): whether the predicted status flag \(\hat{s}_i\) matches the ground-truth status \(s_i\).  

    \item \textbf{Step success rate} (\(\text{Acc}_{\text{step}}\)): a step is considered correct only if all three components (function, arguments, status) are correct simultaneously.  
    \[
    \text{Acc}_{\text{step}} = \frac{1}{N}\sum_{i=1}^{N}\mathbb{1}\{ \hat{f}_i = f_i \wedge \hat{a}_i \equiv a_i \wedge \hat{s}_i = s_i \}.
    \]
\end{itemize}

\paragraph{Performance Comparison.}
\begin{table*}[t]
\centering
\caption{Model performance comparison with screen Visual-only (left) and with screen Visual+A11y (right). Values are reported as percentages.}
\label{tab:action_pred}
\begin{tabular}{lrrrrrrrr}
\toprule
 & \multicolumn{4}{c}{\textbf{Visual-only}} & \multicolumn{4}{c}{\textbf{Visual+A11y}} \\
\cmidrule(lr){2-5} \cmidrule(lr){6-9}
Model & Word & Excel & PowerPoint & Total & Word & Excel & PowerPoint & Total \\
\midrule
GPT-4o & 3.61 & 1.96 & 3.35 & 3.12 & 61.36 & 29.15 & 48.11 & 36.71 \\
GPT-4.1 & 3.60 & 1.88 & 2.55 & 2.82 & 35.46 & 33.13 & 50.98 & 39.19 \\
GPT-o3 & 16.85 & 13.06 & 24.42 & 17.92 & 34.00 & 28.76 & 48.84 & 36.72 \\
GPT-5 & 9.05 & 6.21 & 10.26 & 8.59 & 31.68 & 26.39 & 48.23 & 34.86 \\
Qwen-2.5 7B & 15.70 & 12.75 & 25.09 & 17.52 & 15.64 & 3.56 & 22.51 & 14.18 \\
Qwen-2.5 7B-SFT & 49.10 & 45.12 & 56.53 & 50.08 & 31.68 & 7.44 & 34.99 & 25.78 \\
\bottomrule
\end{tabular}
\end{table*}

Table~\ref{tab:action_pred} reports the results of action prediction under the \emph{visual-only} and \emph{visual+a11y} settings. 
When relying on screenshots alone, all models perform poorly, with accuracy below $20\%$ in most cases. 
This highlights the intrinsic difficulty of inferring precise action arguments purely from pixel-level cues, even for state-of-the-art proprietary VLMs such as GPT-4.1 and GPT-5. 
By contrast, providing accessibility information dramatically boosts performance. For example, GPT-4o improves from $3.12\%$ to $36.71\%$, and GPT-4.1 nearly triples its performance from $2.82\%$ to $39.19\%$. 
This demonstrates that structured element annotations effectively reduce the burden of visual grounding, enabling models to focus on action semantics. 

Furthermore, supervised fine-tuning (SFT) on \name delivers substantial gains. Qwen-2.5~7B improves from $17.52\%$ to $50.08\%$ after SFT in the visual-only setting, a nearly threefold improvement, showing the strong training signal provided by \name. 
However, when a11y information is introduced, the benefits of SFT diminish, suggesting that a11y annotations already encode much of the structural alignment that SFT otherwise learns. 
Overall, these results highlight both the challenge and opportunity presented by \name: action prediction is extremely difficult without explicit structural information, but with accessibility-enhanced input and dataset-driven post-training, models can achieve substantial improvements.

\begin{table}[t]
\centering
\caption{Comparison of Qwen and SFT models with/without A11Y across domains. Each cell shows ``Qwen / SFT'', with bold indicating the better value.}
\label{tab:qwen_sft_a11y}
\resizebox{\textwidth}{!}{
\begin{tabular}{lcccccccc}
\toprule
 & \multicolumn{4}{c}{w/o A11Y} & \multicolumn{4}{c}{w/ A11Y} \\
\cmidrule(lr){2-5} \cmidrule(lr){6-9}
Metric & Excel & Word & PPT & Overall & Excel & Word & PPT & Overall \\
\midrule
Function Match      
 & 61.67 / \textbf{81.07} & 62.97 / \textbf{81.18} & 88.50 / \textbf{91.02} & 69.82 / \textbf{83.93} 
 & 50.61 / \textbf{80.64} & 66.62 / \textbf{83.19} & 90.82 / \textbf{91.53} & 68.95 / \textbf{84.83} \\
Args Match          
 & 12.84 / \textbf{45.43} & 15.80 / \textbf{49.38} & 25.16 / \textbf{56.72} & 17.61 / \textbf{50.34} 
 & 3.60 / \textbf{7.47} & 15.74 / \textbf{31.82} & 22.60 / \textbf{35.16} & 14.25 / \textbf{25.91} \\
Status Match        
 & \textbf{95.96} / 94.13 & \textbf{96.37} / 93.85 & \textbf{97.44} / 96.17 & \textbf{96.56} / 94.59 
 & 85.74 / \textbf{95.27} & \textbf{98.23} / 95.77 & \textbf{99.00} / 96.34 & 94.93 / \textbf{95.79} \\
Args Mismatch Err.  
 & 87.17 / \textbf{54.56} & 84.21 / \textbf{50.62} & 74.84 / \textbf{43.28} & 82.39 / \textbf{49.66} 
 & 96.40 / \textbf{92.53} & 84.26 / \textbf{68.18} & 77.40 / \textbf{64.84} & 85.74 / \textbf{74.10} \\
Coord. OOB          
 & 74.94 / \textbf{63.97} & 66.63 / \textbf{61.39} & 96.15 / \textbf{82.83} & 76.68 / \textbf{67.47} 
 & 74.91 / \textbf{89.77} & 83.58 / \textbf{76.72} & 98.78 / \textbf{90.54} & 84.71 / \textbf{84.73} \\
\bottomrule
\end{tabular}}
\end{table}

\paragraph{Result Analysis.}
To obtain a fine-grained understanding of action prediction, we decompose evaluation into three aspects: (i) \emph{function match}, which tests whether the predicted function type is correct; (ii) \emph{argument match}, which checks whether the predicted arguments (e.g., coordinates or values) align with ground truth; and (iii) \emph{status match}, which verifies whether the model correctly predicts task continuation or completion. In addition, we track two error sources: \emph{Args Mismatch Error}, capturing the proportion of incorrect arguments, and \emph{Coord. OOB}, which reflects out-of-bounds coordinate predictions in the visual-only setting or incorrect element selections when a11y information is available.  

From Table~\ref{tab:qwen_sft_a11y}, we observe that supervised fine-tuning (SFT) brings large gains in \emph{function match} and especially in \emph{argument match}, reducing errors by more than half across domains. The dominant source of failure remains \emph{Args Mismatch Error}, which suggests that grounding actions to the correct interface element is the most challenging aspect. Notably, Coord. OOB contributes significantly to these mismatches, highlighting the model’s difficulty in spatial grounding from raw screenshots.  

Comparing the two evaluation settings, we find that introducing a11y metadata substantially reduces \emph{Coord. OOB} errors, showing that structured semantic cues provide more reliable grounding than relying on visual information alone. However, even with a11y support, argument prediction accuracy lags far behind function prediction, indicating that grounding arguments---either through spatial reasoning or element identification---remains the key bottleneck for reliable GUI action prediction. Overall, these findings highlight that while models can correctly identify the intended operation, achieving precise grounding is still an open challenge, and incorporating structured UI metadata such as a11y is a promising direction.

\section{Conclusion}
In this work, we introduced \name, a large-scale dataset and benchmark suite for advancing research on desktop computer-using agents. \name fills three critical gaps in the field: the lack of realistic task collections, the absence of scalable data collection pipelines, and the shortage of unified benchmarks spanning GUI grounding, screen parsing, and action prediction. Through an automated, LLM-augmented pipeline, we curated over 1.2M steps across thousands of trajectories in widely used Windows applications, paired with rich multimodal annotations including screenshots, accessibility metadata, reasoning traces, and both successful and failed executions. 

Our empirical evaluation highlights both the difficulty of the domain and the promise of \name. State-of-the-art vision–language models exhibit significant limitations when applied out-of-the-box, particularly in grounding and action prediction. Yet, fine-tuning and reinforcement learning on \name deliver consistent improvements, underscoring the dataset’s utility as a training and evaluation resource. Importantly, results remain far from human-level reliability, establishing \name as a challenging but necessary foundation for future progress. 

We release \name, the accompanying benchmark \name-Bench, and the full data collection framework to the research community. We hope these resources catalyze systematic advances in screen understanding, multimodal reasoning, and robust action planning, ultimately bringing practical and reliable computer-using agents closer to reality.

\bibliography{iclr2026_conference}

@article{wang2024mobile,
  title={Mobile-agent: Autonomous multi-modal mobile device agent with visual perception},
  author={Wang, Junyang and Xu, Haiyang and Ye, Jiabo and Yan, Ming and Shen, Weizhou and Zhang, Ji and Huang, Fei and Sang, Jitao},
  journal={arXiv preprint arXiv:2401.16158},
  year={2024}
}

@article{wang2024mobilev2,
  title={Mobile-agent-v2: Mobile device operation assistant with effective navigation via multi-agent collaboration},
  author={Wang, Junyang and Xu, Haiyang and Jia, Haitao and Zhang, Xi and Yan, Ming and Shen, Weizhou and Zhang, Ji and Huang, Fei and Sang, Jitao},
  journal={Advances in Neural Information Processing Systems},
  volume={37},
  pages={2686--2710},
  year={2024}
}

@article{wang2025mobile,
  title={Mobile-agent-e: Self-evolving mobile assistant for complex tasks},
  author={Wang, Zhenhailong and Xu, Haiyang and Wang, Junyang and Zhang, Xi and Yan, Ming and Zhang, Ji and Huang, Fei and Ji, Heng},
  journal={arXiv preprint arXiv:2501.11733},
  year={2025}
}

@article{zheng2024gpt,
  title={Gpt-4v (ision) is a generalist web agent, if grounded},
  author={Zheng, Boyuan and Gou, Boyu and Kil, Jihyung and Sun, Huan and Su, Yu},
  journal={arXiv preprint arXiv:2401.01614},
  year={2024}
}

@article{zhang2024ufo,
  title={Ufo: A ui-focused agent for windows os interaction},
  author={Zhang, Chaoyun and Li, Liqun and He, Shilin and Zhang, Xu and Qiao, Bo and Qin, Si and Ma, Minghua and Kang, Yu and Lin, Qingwei and Rajmohan, Saravan and others},
  journal={arXiv preprint arXiv:2402.07939},
  year={2024}
}

@article{zhang2025ufo2,
  title={Ufo2: The desktop agentos},
  author={Zhang, Chaoyun and Huang, He and Ni, Chiming and Mu, Jian and Qin, Si and He, Shilin and Wang, Lu and Yang, Fangkai and Zhao, Pu and Du, Chao and others},
  journal={arXiv preprint arXiv:2504.14603},
  year={2025}
}

@article{qin2025ui,
  title={Ui-tars: Pioneering automated gui interaction with native agents},
  author={Qin, Yujia and Ye, Yining and Fang, Junjie and Wang, Haoming and Liang, Shihao and Tian, Shizuo and Zhang, Junda and Li, Jiahao and Li, Yunxin and Huang, Shijue and others},
  journal={arXiv preprint arXiv:2501.12326},
  year={2025}
}

@article{zheng2025skillweaver,
  title={Skillweaver: Web agents can self-improve by discovering and honing skills},
  author={Zheng, Boyuan and Fatemi, Michael Y and Jin, Xiaolong and Wang, Zora Zhiruo and Gandhi, Apurva and Song, Yueqi and Gu, Yu and Srinivasa, Jayanth and Liu, Gaowen and Neubig, Graham and others},
  journal={arXiv preprint arXiv:2504.07079},
  year={2025}
}

@article{gur2023real,
  title={A real-world webagent with planning, long context understanding, and program synthesis},
  author={Gur, Izzeddin and Furuta, Hiroki and Huang, Austin and Safdari, Mustafa and Matsuo, Yutaka and Eck, Douglas and Faust, Aleksandra},
  journal={arXiv preprint arXiv:2307.12856},
  year={2023}
}

@article{xie2023openagents,
  title={Openagents: An open platform for language agents in the wild},
  author={Xie, Tianbao and Zhou, Fan and Cheng, Zhoujun and Shi, Peng and Weng, Luoxuan and Liu, Yitao and Hua, Toh Jing and Zhao, Junning and Liu, Qian and Liu, Che and others},
  journal={arXiv preprint arXiv:2310.10634},
  year={2023}
}

@article{cheng2024seeclick,
  title={Seeclick: Harnessing gui grounding for advanced visual gui agents},
  author={Cheng, Kanzhi and Sun, Qiushi and Chu, Yougang and Xu, Fangzhi and Li, Yantao and Zhang, Jianbing and Wu, Zhiyong},
  journal={arXiv preprint arXiv:2401.10935},
  year={2024}
}

@article{lu2024omniparser,
  title={Omniparser for pure vision based gui agent},
  author={Lu, Yadong and Yang, Jianwei and Shen, Yelong and Awadallah, Ahmed},
  journal={arXiv preprint arXiv:2408.00203},
  year={2024}
}

@article{xu2025deskvision,
  title={DeskVision: Large Scale Desktop Region Captioning for Advanced GUI Agents},
  author={Xu, Yibin and Yang, Liang and Chen, Hao and Wang, Hua and Chen, Zhi and Tang, Yaohua},
  journal={arXiv preprint arXiv:2503.11170},
  year={2025}
}

@article{zheng2025vem,
  title={Vem: Environment-free exploration for training gui agent with value environment model},
  author={Zheng, Jiani and Wang, Lu and Yang, Fangkai and Zhang, Chaoyun and Mei, Lingrui and Yin, Wenjie and Lin, Qingwei and Zhang, Dongmei and Rajmohan, Saravan and Zhang, Qi},
  journal={arXiv preprint arXiv:2502.18906},
  year={2025}
}

@article{wang2024officebench,
  title={Officebench: Benchmarking language agents across multiple applications for office automation},
  author={Wang, Zilong and Cui, Yuedong and Zhong, Li and Zhang, Zimin and Yin, Da and Lin, Bill Yuchen and Shang, Jingbo},
  journal={arXiv preprint arXiv:2407.19056},
  year={2024}
}

@article{nayak2025ui,
  title={Ui-vision: A desktop-centric gui benchmark for visual perception and interaction},
  author={Nayak, Shravan and Jian, Xiangru and Lin, Kevin Qinghong and Rodriguez, Juan A and Kalsi, Montek and Awal, Rabiul and Chapados, Nicolas and {\"O}zsu, M Tamer and Agrawal, Aishwarya and Vazquez, David and others},
  journal={arXiv preprint arXiv:2503.15661},
  year={2025}
}

@article{rawles2024androidworld,
  title={Androidworld: A dynamic benchmarking environment for autonomous agents},
  author={Rawles, Christopher and Clinckemaillie, Sarah and Chang, Yifan and Waltz, Jonathan and Lau, Gabrielle and Fair, Marybeth and Li, Alice and Bishop, William and Li, Wei and Campbell-Ajala, Folawiyo and others},
  journal={arXiv preprint arXiv:2405.14573},
  year={2024}
}

@article{zhou2023webarena,
  title={Webarena: A realistic web environment for building autonomous agents},
  author={Zhou, Shuyan and Xu, Frank F and Zhu, Hao and Zhou, Xuhui and Lo, Robert and Sridhar, Abishek and Cheng, Xianyi and Ou, Tianyue and Bisk, Yonatan and Fried, Daniel and others},
  journal={arXiv preprint arXiv:2307.13854},
  year={2023}
}

@article{wu2025gui,
  title={GUI-Actor: Coordinate-Free Visual Grounding for GUI Agents},
  author={Wu, Qianhui and Cheng, Kanzhi and Yang, Rui and Zhang, Chaoyun and Yang, Jianwei and Jiang, Huiqiang and Mu, Jian and Peng, Baolin and Qiao, Bo and Tan, Reuben and others},
  journal={arXiv preprint arXiv:2506.03143},
  year={2025}
}

@article{sun2024genesis,
  title={Os-genesis: Automating gui agent trajectory construction via reverse task synthesis},
  author={Sun, Qiushi and Cheng, Kanzhi and Ding, Zichen and Jin, Chuanyang and Wang, Yian and Xu, Fangzhi and Wu, Zhenyu and Jia, Chengyou and Chen, Liheng and Liu, Zhoumianze and others},
  journal={arXiv preprint arXiv:2412.19723},
  year={2024}
}

@article{wang2024large,
  title={Large action models: From inception to implementation},
  author={Wang, Lu and Yang, Fangkai and Zhang, Chaoyun and Lu, Junting and Qian, Jiaxu and He, Shilin and Zhao, Pu and Qiao, Bo and Huang, Ray and Qin, Si and others},
  journal={arXiv preprint arXiv:2412.10047},
  year={2024}
}

@article{gu2024survey,
  title={A survey on llm-as-a-judge},
  author={Gu, Jiawei and Jiang, Xuhui and Shi, Zhichao and Tan, Hexiang and Zhai, Xuehao and Xu, Chengjin and Li, Wei and Shen, Yinghan and Ma, Shengjie and Liu, Honghao and others},
  journal={arXiv preprint arXiv:2411.15594},
  year={2024}
}

@article{hou2025model,
  title={Model context protocol (mcp): Landscape, security threats, and future research directions},
  author={Hou, Xinyi and Zhao, Yanjie and Wang, Shenao and Wang, Haoyu},
  journal={arXiv preprint arXiv:2503.23278},
  year={2025}
}

@inproceedings{yao2023react,
  title={React: Synergizing reasoning and acting in language models},
  author={Yao, Shunyu and Zhao, Jeffrey and Yu, Dian and Du, Nan and Shafran, Izhak and Narasimhan, Karthik and Cao, Yuan},
  booktitle={International Conference on Learning Representations (ICLR)},
  year={2023}
}

@article{haverty2005new,
  title={New accessibility model for Microsoft Windows and cross platform development},
  author={Haverty, Rob},
  journal={ACM SIGACCESS Accessibility and Computing},
  number={82},
  pages={11--17},
  year={2005},
  publisher={ACM New York, NY, USA}
}

@article{yang2023set,
  title={Set-of-mark prompting unleashes extraordinary visual grounding in gpt-4v},
  author={Yang, Jianwei and Zhang, Hao and Li, Feng and Zou, Xueyan and Li, Chunyuan and Gao, Jianfeng},
  journal={arXiv preprint arXiv:2310.11441},
  year={2023}
}

@article{deng2023mind2web,
  title={Mind2web: Towards a generalist agent for the web},
  author={Deng, Xiang and Gu, Yu and Zheng, Boyuan and Chen, Shijie and Stevens, Sam and Wang, Boshi and Sun, Huan and Su, Yu},
  journal={Advances in Neural Information Processing Systems},
  volume={36},
  pages={28091--28114},
  year={2023}
}

@article{li2025screenspot,
  title={Screenspot-pro: Gui grounding for professional high-resolution computer use},
  author={Li, Kaixin and Meng, Ziyang and Lin, Hongzhan and Luo, Ziyang and Tian, Yuchen and Ma, Jing and Huang, Zhiyong and Chua, Tat-Seng},
  journal={arXiv preprint arXiv:2504.07981},
  year={2025}
}

@article{rawles2023androidinthewild,
  title={Androidinthewild: A large-scale dataset for android device control},
  author={Rawles, Christopher and Li, Alice and Rodriguez, Daniel and Riva, Oriana and Lillicrap, Timothy},
  journal={Advances in Neural Information Processing Systems},
  volume={36},
  pages={59708--59728},
  year={2023}
}

@article{cua2025,
  title={Computer-Using Agent: Introducing a universal interface for AI to interact with the digital world},
  author={OpenAI},
  year={2025},
  url={https://openai.com/index/computer-using-agent},
}

@article{zhang2024large,
  title={Large language model-brained gui agents: A survey},
  author={Zhang, Chaoyun and He, Shilin and Qian, Jiaxu and Li, Bowen and Li, Liqun and Qin, Si and Kang, Yu and Ma, Minghua and Liu, Guyue and Lin, Qingwei and others},
  journal={arXiv preprint arXiv:2411.18279},
  year={2024}
}

@inproceedings{zhang2025api,
  title={API Agents vs. GUI Agents: Divergence and Convergence},
  author={Zhang, Chaoyun and He, Shilin and Li, Liqun and Qin, Si and Kang, Yu and Lin, Qingwei and Rajmohan, Saravan and Zhang, Dongmei},
  booktitle={ICML 2025 Workshop on Computer Use Agents}
}

@article{bai2025qwen2,
  title={Qwen2. 5-vl technical report},
  author={Bai, Shuai and Chen, Keqin and Liu, Xuejing and Wang, Jialin and Ge, Wenbin and Song, Sibo and Dang, Kai and Wang, Peng and Wang, Shijie and Tang, Jun and others},
  journal={arXiv preprint arXiv:2502.13923},
  year={2025}
}

@article{gou2024navigating,
  title={Navigating the digital world as humans do: Universal visual grounding for gui agents},
  author={Gou, Boyu and Wang, Ruohan and Zheng, Boyuan and Xie, Yanan and Chang, Cheng and Shu, Yiheng and Sun, Huan and Su, Yu},
  journal={arXiv preprint arXiv:2410.05243},
  year={2024}
}

@article{xu2024aguvis,
  title={Aguvis: Unified pure vision agents for autonomous gui interaction},
  author={Xu, Yiheng and Wang, Zekun and Wang, Junli and Lu, Dunjie and Xie, Tianbao and Saha, Amrita and Sahoo, Doyen and Yu, Tao and Xiong, Caiming},
  journal={arXiv preprint arXiv:2412.04454},
  year={2024}
}

@article{hurst2024gpt,
  title={Gpt-4o system card},
  author={Hurst, Aaron and Lerer, Adam and Goucher, Adam P and Perelman, Adam and Ramesh, Aditya and Clark, Aidan and Ostrow, AJ and Welihinda, Akila and Hayes, Alan and Radford, Alec and others},
  journal={arXiv preprint arXiv:2410.21276},
  year={2024}
}

@article{openai2025gpt4_1,
  author    = {OpenAI},
  title     = {Introducing GPT-4.1 in the API},
  year      = {2025},
  month     = apr,
  day       = {14},
  url       = {https://openai.com/index/gpt-4-1/},
  note      = {Accessed: 2025-09-18}
}

@misc{openai2025o3_o4mini,
  author       = {OpenAI},
  title        = {Introducing OpenAI o3 and o4-mini},
  howpublished = {\url{https://openai.com/index/introducing-o3-and-o4-mini/}},
  year         = {2025},
  month        = apr,
  day          = {16},
  note         = {Accessed: 2025-09-18}
}

@misc{openai2025gpt5,
  author       = {OpenAI},
  title        = {GPT-5 is here},
  howpublished = {\url{https://openai.com/gpt-5/}},
  year         = {2025},
  note         = {Accessed: 2025-09-18}
}
\bibliographystyle{iclr2026_conference}

\appendix

\section{Action Set}
To support diverse interaction across desktop applications, we design a unified action set that combines general-purpose GUI operations with application-specific APIs using MCP servers. The action set is deliberately lightweight yet expressive, enabling agents to cover the full spectrum of common productivity tasks while remaining tractable for model training.

The GUI actions (\texttt{click}, \texttt{type}, \texttt{drag}, \texttt{wheel\_mouse\_input}) form the foundation of interaction, as they are applicable to any graphical user interface. These actions abstract low-level mouse and keyboard events into structured calls, supporting variants such as absolute and normalized coordinates, modifier keys, and multi-step operations.

On top of this universal layer, we extend the action set with fine-grained APIs for \textbf{Word}, \textbf{Excel}, and \textbf{PowerPoint}. These APIs expose high-level document, spreadsheet, and presentation semantics, such as inserting tables, modifying cell values, reordering columns, adjusting font properties, or setting slide backgrounds. By combining GUI-agnostic operations with domain-specific APIs, the action set achieves both generality and efficiency: agents can rely on GUI actions for arbitrary interfaces while exploiting APIs for structured tasks where precise semantics matter.

Table~\ref{tab:action_set} summarizes the full action set. This unified design allows agents trained on \name to operate seamlessly across heterogeneous applications, balancing robustness with expressivity.

\begin{table}[t]
\centering
\small
\caption{Supported actions across Word, Excel, and PowerPoint, grouped by shared GUI actions and application-specific APIs.}
\label{tab:action_set}
\begin{tabular}{p{3cm} p{8cm} p{2cm}}
\toprule
\textbf{Action} & \textbf{Description} & \textbf{Type} \\
\midrule
click & Click at a given position (absolute or normalized), supporting left/right/middle/x button, single/double click, and optional modifier key. & GUI \\
type & Type text or hotkeys at a position, with options for clearing text or focusing on control. Supports special keys like \{VK\_CONTROL\}c. & GUI \\
drag & Drag from a start to an end position with configurable mouse button, duration, and optional key hold (e.g., shift, control). & GUI \\
wheel\_mouse\_input & Scroll at a given position with positive (up) or negative (down) wheel distance. & GUI \\
\midrule
insert\_table & Insert a table with a specified number of rows and columns. & Word API \\
select\_text & Select exact text in the document. & Word API \\
select\_table & Select a table by its index number. & Word API \\
select\_paragraph & Select paragraphs by start and end indices, with option to restrict to non-empty ones. & Word API \\
save\_as & Save the document with specified directory, file name, and extension (default: PDF). & Word API \\
set\_font & Change font family and/or size. & Word API \\
\midrule
table2markdown & Extract the contents of a worksheet table into Markdown format. & Excel API \\
insert\_excel\_table & Insert a table (list of lists) into a sheet at a specified starting cell. & Excel API \\
select\_table\_range & Select a range of cells by coordinates in a sheet. & Excel API \\
set\_cell\_value & Set the value (or formula) of a specific cell. & Excel API \\
auto\_fill & Autofill values in a specified cell range. & Excel API \\
reorder\_columns & Reorder columns in a sheet according to a given list of column names. & Excel API \\
\midrule
set\_background\_color & Change the slide background color using a hex RGB value, for selected or all slides. & PowerPoint API \\
save\_as & Save the presentation with specified directory, file name, and extension (default: PowerPointX). Optionally save slides as images. & PowerPoint API \\
\bottomrule
\end{tabular}
\end{table}

\section{\name Schema}
Each execution step in \name is stored as a structured JSON object following a unified schema. 
This schema ensures consistency across tasks and provides rich multimodal supervision for grounding, parsing, and action prediction. 
Table~\ref{tab:schema} summarizes the key fields.

\begin{table}[h]
\centering
\caption{Execution Step Schema for \name. Each entry records metadata, screenshots, accessibility data, reasoning traces, and actions.}
\label{tab:schema}
\renewcommand{\arraystretch}{1.2}
\begin{tabular}{p{4cm}| p{9cm}}
\toprule
\textbf{Field} & \textbf{Description} \\
\midrule
\texttt{execution\_id} & Unique identifier for the execution instance (e.g., \texttt{word\_1\_1}). \\\midrule
\texttt{app\_domain} & Application domain (e.g., Word, Excel, PowerPoint). \\\midrule
\texttt{request} & Natural-language task description provided to the agent. \\\midrule
\texttt{template} & Environment template file used for instantiation. \\\midrule
\texttt{step\_id} / \texttt{total\_steps} & Current step index and total number of steps in the trajectory. \\\midrule
\texttt{evaluation} & Automatic assessment of the step, including reasoning, evidence, sub-scores, and a final completeness label. \\\midrule
\texttt{step.screenshots} & Multiple synchronized screenshots: clean view, full desktop, annotated version, and selected-controls view. \\\midrule
\texttt{step.ui\_tree} & Hierarchical UI structure with element IDs, names, control types, bounding boxes, and children. \\\midrule
\texttt{step.control\_infos} & Metadata from accessibility APIs and merged control sources, providing bounding boxes, labels, and semantic text. \\\midrule
\texttt{step.observation} & Agent’s textual observation of the current state. \\\midrule
\texttt{step.thought} & Agent’s intermediate reasoning for the next action. \\\midrule
\texttt{step.action} & Executed action, including function type (e.g., \texttt{click}), arguments, target coordinates, and status flag. \\\midrule
\texttt{status} & Overall status of the step (\texttt{CONTINUE} or \texttt{FINISH}). \\
\bottomrule
\end{tabular}
\end{table}

\paragraph{Discussion.}
The schema integrates three complementary perspectives: 
(i) \textbf{Visual context} through multi-view screenshots, 
(ii) \textbf{Structural context} via accessibility metadata and hierarchical UI trees, and 
(iii) \textbf{Cognitive traces} through observations, reasoning, actions, and evaluations. 
This rich structure allows \name to jointly support grounding, parsing, and action prediction, while enabling both supervised training and fine-grained evaluation.
By standardizing every execution step, the schema provides a scalable foundation for reproducibility and extensibility across applications.

\section{Baseline Details}
\label{sec:baseline_detail}
We summarize the baselines evaluated on \name across the three core tasks: 
GUI grounding, screen parsing, and action prediction. 
These baselines include both general-purpose vision--language models and domain-specific approaches designed for GUI reasoning. 
Below we briefly introduce each group of models.

\subsection{GUI Grounding}
\begin{itemize}[leftmargin=*,itemsep=2pt]
  \item \textbf{GPT-4o}~\cite{hurst2024gpt}: proprietary multimodal VLM used off-the-shelf for grounding.
  \item \textbf{GPT-4.1}~\cite{openai2025gpt4_1}: proprietary VLM emphasizing instruction-following and tool use.
  \item \textbf{o3}~\cite{openai2025o3_o4mini}: OpenAI “reasoning” model family; evaluated zero-shot for grounding.
  \item \textbf{GPT-5}~\cite{openai2025gpt5}: latest OpenAI flagship; strong general reasoning baseline.
  \item \textbf{Qwen2.5-VL-7B}~\cite{bai2025qwen2}: open-source 7B multimodal baseline.
  \item \textbf{UGround-7B}~\cite{gou2024navigating}: GUI visual grounding model (Qwen2-VL backbone) trained with large-scale GUI data.
  \item \textbf{Aguvis-7B}~\cite{xu2024aguvis}: vision-centric GUI agent with a unified cross-platform action space.
  \item \textbf{UI-TARS-1.5 (7B)}~\cite{qin2025ui}: multimodal agent optimized for GUI reasoning and interactive tasks.
  \item \textbf{GUI-Actor (7B)}~\cite{wu2025gui}: coordinate-free grounding model with an attention-based action head.
  \item \textbf{SFT variants}: Qwen2.5-VL-7B fine-tuned on \name for task-adapted grounding.
\end{itemize}

\subsection{Screen Parsing}
\begin{itemize}[leftmargin=*,itemsep=2pt]
  \item \textbf{GPT-4o / GPT-4.1 / o3 / GPT-5}~\cite{hurst2024gpt,openai2025gpt4_1,openai2025o3_o4mini,openai2025gpt5}: general-purpose VLMs used off-the-shelf for detection/localization.
  \item \textbf{Qwen2.5-VL-7B}~\cite{bai2025qwen2}: open-source multimodal baseline.
  \item \textbf{OmniParser / OmniParser-v2}~\cite{lu2024omniparser}: screen-parsing tools that produce element sets (names + bounding boxes) from raw screenshots.
\end{itemize}

\subsection{Action Prediction}
\begin{itemize}[leftmargin=*,itemsep=2pt]
  \item \textbf{GPT-4o / GPT-4.1 / o3 / GPT-5}~\cite{hurst2024gpt,openai2025gpt4_1,openai2025o3_o4mini,openai2025gpt5}: proprietary VLMs evaluated in both \emph{visual-only} and \emph{visual+a11y} settings.
  \item \textbf{Qwen2.5-VL-7B}~\cite{bai2025qwen2}: open-source baseline for structured action generation.
  \item \textbf{Qwen2.5-VL-7B-SFT}: supervised fine-tuning on \name for step-wise function/argument/status prediction.
\end{itemize}

\paragraph{Summary.} 
Together, these baselines span a spectrum from general-purpose VLMs to specialized GUI-focused agents. 
This diversity allows us to systematically evaluate the unique challenges posed by \name across grounding, parsing, and action prediction, 
and to measure how far current models remain from robust, human-level computer-using agents.

\section{Implementation Details}
\label{sec:implement}

\paragraph{Data Collection.} 
To construct \name, we deployed a cluster of \textbf{15 Windows 11 virtual machines}, each provisioned with 4 CPU cores. 
These VMs executed tasks in parallel, enabling efficient large-scale trajectory collection. 
Task execution followed a two-phase strategy: in \textbf{Phase 1}, GPT-4o was used as the agent; in \textbf{Phase 2}, all failed tasks were re-executed with GPT-4.1 for recovery (see Section~\ref{sec:method}). 
Both models were queried with the temperature fixed at \texttt{0.0} to ensure deterministic outputs and reproducibility.

\paragraph{Model Access.} 
For evaluation on \name-Bench, we used multiple OpenAI models, including GPT-4o, GPT-4.1, o3, and GPT-5, all accessed via the \textbf{Azure OpenAI Service}. 
Unless otherwise stated, the decoding temperature was set to \texttt{0.0} across all experiments to minimize variance and ensure consistent evaluation.

\paragraph{Fine-tuning and Training.} 
All supervised fine-tuning (SFT) and reinforcement learning (RL) experiments were conducted on a compute cluster equipped with \textbf{NVIDIA A100 GPUs} (40GB memory per GPU). 
Specifically, each run was distributed across 4 A100 GPUs using mixed-precision training (FP16) for efficiency. 
We adopted standard optimization settings following prior work on multimodal fine-tuning, with learning rates tuned over \(\{1e\text{-}5, 5e\text{-}6, 1e\text{-}6\}\). 
Checkpointing and gradient accumulation were applied to ensure stable training for long trajectories. 

\section*{Ethics Statement}

This work presents \name, a large-scale dataset and benchmark for GUI agents on Windows applications. We carefully considered ethical implications throughout the data collection, processing, and release pipeline. First, no human subjects were directly involved in the data collection process, and thus no personally identifiable information (PII) or sensitive user data is included. All queries and trajectories were generated and executed within controlled sandbox environments to ensure both privacy and security. Second, we adhered to software license terms and platform usage guidelines, ensuring that the collection process does not violate proprietary restrictions or legal compliance requirements. Third, we performed multiple post-processing stages, including trajectory validation, data sanitization, and structuring, to filter out incomplete, low-quality, or potentially misleading samples, thereby reducing the risk of harmful insights or erroneous model behaviors.

We acknowledge the potential downstream misuse of GUI automation technologies, such as unauthorized system manipulation or exploitation of accessibility features. To mitigate this, we restrict dataset release to non-sensitive application contexts (Word, Excel, PowerPoint) and exclude scenarios that could pose privacy, security, or safety risks. The dataset is intended solely for academic research on improving robustness, generalization, and evaluation of GUI agents. All models and baselines are evaluated under responsible-use guidelines, and we encourage future researchers to follow the same principles.

\section*{Reproducibility Statement}

We place strong emphasis on reproducibility. To this end, we will release all data collection code, execution framework, and templates used to instantiate user queries. The full dataset \name, along with the benchmark split \name-Bench, will also be publicly available. Detailed descriptions of the pipeline are provided in Section~\ref{sec:method}, with task definitions summarized in Table~\ref{tab:task_io}, filtering statistics in Table~\ref{tab:task_filtering_stats}, and execution details in Appendix~\ref{sec:implement}. Additional implementation details, hyperparameters, and evaluation protocols are included in the supplementary material. Together, these resources ensure that both our dataset creation and experimental results can be fully reproduced, verified, and extended by the research community.

\section*{LLM Usage Statement}
In preparing this work, we used large language models (LLMs) strictly as an assistive tool for text polishing and minor language refinement. All research ideas, technical designs, analyses, and conclusions were conceived and carried out entirely by the authors.

\end{document}